\documentclass[sigplan,10pt]{acmart}
\usepackage{amsmath,amsfonts}
\usepackage{algorithmic}
\usepackage{graphicx}
\usepackage{textcomp}
\usepackage{xcolor}
\usepackage{multirow}

\usepackage{pgfplotstable}
\usepackage{pgfplots}
\usepackage{caption}
\usepgfplotslibrary{groupplots}
\usepackage{tikz}

\usepackage{pifont}
\usepackage{color, colortbl}
\definecolor{darkgreen}{rgb}{0,0.5,0}
\definecolor{dgreen}{rgb}{0.7,1,0.7}
\definecolor{lgreen}{rgb}{0.5,1,0.5}
\definecolor{llgreen}{rgb}{0.9,1,0.9}
\definecolor{green}{rgb}{0,1,0}
\definecolor{yellow}{rgb}{1,1,0}
\definecolor{lyellow}{rgb}{1,1,0.75}
\definecolor{red}{rgb}{1,0,0}
\definecolor{lred}{rgb}{1,0.85,0.85}
\definecolor{blue}{rgb}{0,0,1} 
\definecolor{lblue}{rgb}{0.3,0.3,1.0}
\definecolor{llblue}{rgb}{0.8,0.8,1.0}
\definecolor{gray}{gray}{0.50}
\definecolor{gray1}{gray}{0.40}
\definecolor{lgray}{gray}{0.75}
\definecolor{llgray}{gray}{0.95}
\definecolor{applegreen}{rgb}{0.55, 0.71, 0.0}
\definecolor{lightgreen}{rgb}{0.56, 0.93, 0.56}
\definecolor{palegreen}{rgb}{0.85, 1, 0.85}

\usepackage[linesnumbered,ruled,vlined]{algorithm2e}

\usepackage{hyperref}
\usepackage{cleveref}
\usepackage{listings}
\definecolor{mygreen}{rgb}{0,0.6,0}
\definecolor{mygray}{rgb}{0.5,0.5,0.5}
\definecolor{mymauve}{rgb}{0.58,0,0.82}

\lstdefinestyle{customc}{
  belowcaptionskip=1\baselineskip,
  breaklines=true,
  xleftmargin=1em,                % let compare the number list in the paragraph (added by Andrea)
  xrightmargin=1em,               % let compare the number list in the paragraph (added by Andrea)
  language=c,
  numbers=left,
  firstnumber=1,
  emph={int,char,double,float,unsigned,__float128,half,bfloat16,quad,FlexFloat,flytes,Flytes,mpfr_t,vpfloat},
  emphstyle={\color{green!40!black}},
  numberfirstline=true,
  showstringspaces=false,
  frame=single,
  basicstyle=\ttfamily\lst@ifdisplaystyle\footnotesize\fi,
  keywordstyle=\bfseries\color{green!40!black},
  commentstyle=\itshape\color{purple!40!black},
  identifierstyle=\color{blue!40!black},
  stringstyle=\color{orange},
  morecomment=[l]{//},
}

\renewcommand\footnotetextcopyrightpermission[1]{} % no copyright note
\acmConference[arXiv]{}{Nov}{2025.}
\acmBooktitle{}
\acmPrice{}
\acmISBN{}
\acmDOI{}
\makeatletter
\let\@acmConference@name\@empty
\let\@acmConference@shortname\@empty
\makeatother

\begin{document}
% Below should prevent text going beyond column width:
\emergencystretch 3em

\title{Library Liberation: Competitive Performance Matmul Through Compiler-composed Nanokernels}

% \affiliation{%

% \affiliation{%

% \author{Arun Thangamani, Md Asghar Ahmad Shahid, Adam Siemieniuk, Rolf Morel, Renato Golin, and Alexander Heinecke}
% \affiliation{%
%   \institution{Intel Corporation}
%   \city{}
%   \country{}
% }

\author{Arun Thangamani}
\affiliation{%
 \institution{Intel Advanced Technologies Group,\\Intel Corporation, India.}
 \city{}
 \state{}
 \country{}}
 %\email{larst@affiliation.org}

\author{Md Asghar Ahmad Shahid}
\affiliation{%
 \institution{Intel Advanced Technologies Group,\\Intel Corporation, India.}
 \city{}
 \state{}
 \country{}}
 
 \author{Adam Siemieniuk}
\affiliation{%
 \institution{Intel Advanced Technologies Group,\\Intel Corporation, Switzerland.}
 \city{}
 \state{}
 \country{}}
 
 \author{Rolf Morel}
\affiliation{%
 \institution{Intel Advanced Technologies Group,\\Intel Corporation, UK.}
 \city{}
 \state{}
 \country{}}
 %\email{larst@affiliation.org}
 
 \author{Renato Golin}
\affiliation{%
 \institution{Intel Advanced Technologies Group,\\Intel Corporation, UK.}
 \city{}
 \state{}
 \country{}}
 
 \author{Alexander Heinecke}
\affiliation{%
 \institution{Intel Advanced Technologies Group,\\Intel Corporation, USA.}
 \city{}
 \state{}
 \country{}}

\begin{abstract}
% The rapidly evolving landscape of AI and machine learning workloads has widened the semantic gap between high-level domain operations and the efficient utilization of modern processors. Achieving near-peak performance often requires deep hardware expertise: experts either handcraft target-specific kernels (e.g., DeepSeek) or build specialized libraries (e.g., CUTLASS). For most ML practitioners, the former is infeasible and the latter introduces additional dependencies and complexity.

% This paper presents a compilation scheme that automatically generates scalable, high-performance microkernels by leveraging the MLIR vector dialect to bridge the gap between domain-specific operations and processor capabilities. Our approach eliminates much of the need for low-level libraries by enabling the compiler itself to generate near-optimal code. Central to this design is the ability to select and compose nanokernels low-level IR constructs with near-optimal register utilization into efficient microkernels.

% We implement this approach in an MLIR-based compiler targeting both vector- and tile-based CPU instructions. Our experiments demonstrate that the generated nanokernels are production-quality, scalable, and competitive with state-of-the-art microkernel libraries.
The rapidly evolving landscape of AI and machine learning workloads has widened the gap between high-level domain-specific operations and efficient hardware utilization. Achieving near-peak performance still demands deep hardware expertise—experts either handcraft target-specific kernels (e.g., DeepSeek) or rely on specialized libraries (e.g., CUTLASS)—both of which add complexity and limit scalability for most ML practitioners.

This paper introduces a compilation scheme that automatically generates scalable, high-performance microkernels by leveraging multi-level lowering to bridge the gap between domain-level operations and processor-specific capabilities. Our approach removes dependence on low-level libraries by enabling the compiler itself to auto-generate near-optimal code directly. At its core is a mechanism for composing nanokernels from low-level IR constructs with near-optimal register utilization, forming efficient microkernels tailored to each target.
We implement this technique in an MLIR-based compiler supporting both vector- and tile-based CPU instructions. Experiments show that the MLIR generated nanokernels are of production-quality, and competitive with state-of-the-art microkernel libraries.
\end{abstract}




\maketitle

\section{Introduction and Motivation} 
\label{sec:introduction}

Achieving near-peak hardware performance from generic code remains notoriously challenging. Even in machine learning (ML), where programs are dominated and highly structured by a small set of key operations, attaining peak efficiency still requires deep hardware expertise. Despite significant advances in compiler technology, the gap between generic compilation and expert-tuned performance persists.
ML workloads fundamentally rely on two computational primitives: contractions (e.g., tensor and matrix multiplications) and elementwise operations~\cite{mlperf-infer, mlperf-train, fathom}. While contractions can achieve high computational intensity in isolation, the execution of neural networks often exposes a critical bottleneck: the memory wall between consecutive operations. To mitigate this, frameworks increasingly employ operation fusion~\cite{dl-comp-survey, onednn, flashattention}, reducing memory traffic and improving throughput.

Automatic performance optimization has long been a central goal of modern compilers. General-purpose infrastructures such as LLVM and TVM~\cite{llvmir, tvm} effectively handle a wide range of transformations, including code simplification, strength reduction, and data layout optimization. However, achieving strong loop vectorization—essential for compute-intensive operations like General Matrix Multiply (GEMM)—has proven difficult to automate using purely mathematical methods such as polyhedral analysis~\cite{polyhedral}. To overcome these limitations, developers often rely on domain-specific languages (DSLs)~\cite{triton, tensorcomprehensions} or vendor-optimized libraries~\cite{libxsmm, pytorch}. While these solutions deliver high performance, they frequently lag in supporting emerging hardware features and novel operator patterns~\cite{flashattention3, fast-alg-cnn}.

Several frameworks, including Triton~\cite{triton}, OpenCL~\cite{opencl}, and SYCL~\cite{sycl}, aim to bridge the gap between productivity and performance by exposing low-level abstractions that allow developers to handcraft high-performance kernels. Although these approaches differ in portability, simplicity, and efficiency, they all still require substantial hardware-specific expertise. An alternative strategy is to leverage microkernels~\cite{goto2008anatomy} provided through vendor-tuned libraries such as CUTLASS~\cite{cutlass} and OneDNN~\cite{onednn}. Some of these libraries go further by employing runtime JIT compilation to specialize microkernels for specific workloads, as exemplified by the Tensor Processing Primitives (TPP)~\cite{tpp} implemented in \texttt{libxsmm}~\cite{libxsmm}. While effective, such designs introduce additional complexities, including runtime overheads, specialized dependencies, and challenges in maintaining compiler-library alignment.

Recent efforts such as TPP-MLIR~\cite{tppmlir} address these challenges by integrating high-level transformations for linear algebra operators directly into the compiler. These frameworks map contraction operations to optimized microkernel calls, achieving up to 90\% of peak CPU performance out-of-the-box. However, they still depend on external libraries and JIT-based code generation to achieve optimal efficiency.

In this work, we propose embedding expert library knowledge directly into the compiler, eliminating the reliance on external libraries and runtime specialization. Our approach moves target-aware code generation—traditionally confined to external libraries such as \texttt{libxsmm}—into the MLIR~\cite{mlir} compiler framework itself. By leveraging the MLIR {\tt vector} dialect as a target-agnostic entry point, we lower contraction operations through target-specific dialects into optimized LLVM IR~\cite{llvmir}. The resulting code achieves performance comparable to, and often exceeding, state-of-the-art microkernel libraries, while providing a reusable, extensible, and sustainable compilation foundation for future ML workloads.
The key novelty of this work lies in demonstrating that a compiler can automatically generate highly optimized code (name it as nanokernels)—achieving library-level performance—without relying on external JIT-based or library-dependent implementations. To the best of our knowledge, no existing compiler framework has tried and attained this level of performance through fully compiler-generated nanokernels.
%
% The main contributions of this paper are as follows:
% \begin{itemize}
% \item[1.] \textbf{Design.} We present a compilation scheme for generating target-specific, register-tiled GEMM nanokernels, leveraging upstream MLIR dialects and transformations within the TPP-MLIR compiler. We also plan to upstream this compilation scheme into the LLVM/MLIR compiler infrastructure.
% \item[2.] \textbf{Code generation.} We propose a code generation technique that lowers contraction operations directly to target-specific nanokernels. This approach eliminates the need for external microkernel libraries and runtime JIT compilation.
% \item[3.] \textbf{Implementation and evaluation.} We implement our nanokernel generation technique in the TPP-MLIR compiler and evaluate its performance against state-of-the-art microkernel libraries on recent Intel processors.
% \item[4.] \textbf{Integration.} We simplify the overall TPP-MLIR compilation flow through nanokernel generation, enabling a unified compilation path for GEMMs within the MLIR/LLVM compiler infrastructure.
% \end{itemize} 
%
The main contributions of this paper are: \\% as follows:\\
%\begin{itemize}
\noindent \textbf{1. Design.} We present a compilation scheme for generating target-specific, register-tiled GEMM nanokernels, leveraging upstream MLIR dialects and transformations within the TPP-MLIR compiler. We also plan to upstream this compilation scheme into the LLVM/MLIR compiler infrastructure.

\noindent \textbf{2. Code generation.} We propose a code generation technique that lowers contraction operations directly to target-specific nanokernels. This approach eliminates the need for external microkernel libraries and runtime JIT compilation.

\noindent \textbf{3. Implementation and evaluation.} We implement our nanokernel generation technique in the TPP-MLIR compiler and evaluate its performance against state-of-the-art microkernel libraries on recent Intel processors.

\noindent \textbf{4. Integration.} We simplify the overall TPP-MLIR compilation flow through nanokernel generation, enabling a unified compilation path for GEMMs using the MLIR/LLVM compiler infrastructure.

\section{Background}
\label{sec:background}
% \subsection{Batch-Reduce GEMM}
% The BRGEMM microkernel~\cite{brgemm} involves executing multiple \textit{tile} GEMM kernels in a parallel (batch) fashion and accumulating them all onto the same tile (reduce).
% The source of the tiles need not be contiguous or non-overlapping, allowing one to implement various algorithms such as dot-product, outer-product, and convolution. The key gains with this algorithm for our work are: (1) to keep the accumulation tile in registers for all partial GEMMs, and (2) to fuse the followup element-wise operations at the end of the BRGEMM loop.

\subsection{GEMM to Batch-Reduce GEMM}
% Matrix multiplication (GEMM) is a fundamental dense linear algebra operation used throughout machine learning and scientific computing. A simple matmul operation can be represented in MLIR using the {\tt linalg.matmul}. In many machine learning workloads—particularly in transformer models, convolution lowering, and attention mechanisms—multiple small matrix multiplications must be accumulated into the same output tensor. Naively, this pattern can be expressed as a loop over a batch dimension, where each batch slice contributes to the final output.

% The BRGEMM ({\tt linalg.batch\_reduce\_matmul})~\cite{brgemm} is such a pattern that runs multiple \textit{tile} GEMM kernels in a batched fashion, accumulating their results into the same tile (reduce). Importantly, the source tiles are not required to be contiguous or non-overlapping, which enables the implementation of many algorithms such as dot-product, outer-product, and convolution. For our work, BRGEMM provides two key advantages:
% (1) the accumulation tile can remain in registers across all partial GEMMs, reducing memory traffic, and
% (2) follow-up element-wise operations can be naturally fused at the end of the BRGEMM loop, improving efficiency.

Matrix multiplication (GEMM) is a fundamental dense linear algebra operation that underpins a wide range of machine learning and scientific computing workloads.
In MLIR, it can be represented using the {\tt linalg.matmul} operation.
In modern workloads—such as transformer models, convolution lowering, and attention mechanisms—multiple small matrix multiplications are often accumulated into a common output tensor, typically expressed as a loop over a batch dimension.

The Batch-Reduce GEMM~\cite{brgemm} ({\tt linalg.contract} or {\tt linalg.batch\_reduce\_matmul} or {\tt linalg.generic} in MLIR) captures this pattern explicitly by performing multiple tile-level GEMMs in a batched fashion and accumulating results into the same output tile through a reduction step.
The source tiles need not be contiguous, enabling efficient implementations of diverse patterns such as dot products, outer products, and convolutions.
For our work, BRGEMM offers two main advantages:
(1)~the accumulation tile remains resident in registers across partial GEMMs, reducing memory traffic; and
(2)~element-wise operations can be naturally fused at the end of the BRGEMM loop, improving data locality and execution efficiency.

\usetikzlibrary{matrix,positioning}
\begin{figure}[t]
    \centering
    \begin{tikzpicture}
        % First matrix
        \matrix (A) [matrix of math nodes,left delimiter={[},right delimiter={]}] {
            0 & 1 & 2 & 3 \\
            4 & 5 & 6 & 7 \\
            8 & 9 & 10 & 11 \\
            12 & 13 & 14 & 15 \\
        };
        \node[below=-1pt of A.south] {(a) Flat 4$\times$4 matrix};

        % Second matrix
        \matrix (B) [matrix of math nodes,right=0.5cm of A,right delimiter={]},left delimiter={[}] {
            (0, 4) & (1, 5)  & (2,6) & (3,7) \\
            (8, 12) & (9, 13)  & (10, 14) & (11, 15) \\
        };
        \node[below=9pt of B.south] {(b) VNNI:2 packed 2$\times$4$\times$2 matrix};
    \end{tikzpicture}
    \vspace{-.75cm}
    %\caption{Packing a 4$\times$4 row-major order matrix(a) to VNNI(b)}
    \caption{Packing a 4$\times$4 row-major order matrix to VNNI}
    \label{fig:vnni-packing}
    \vspace{-.45cm}
\end{figure}
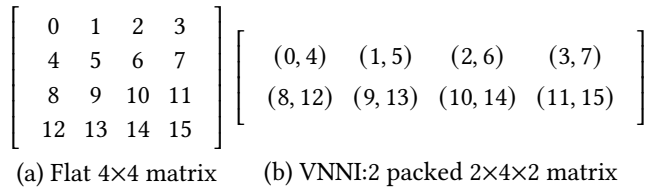

\begin{figure*}%[t]
\centering
\resizebox{.99\textwidth}{!}{ \input{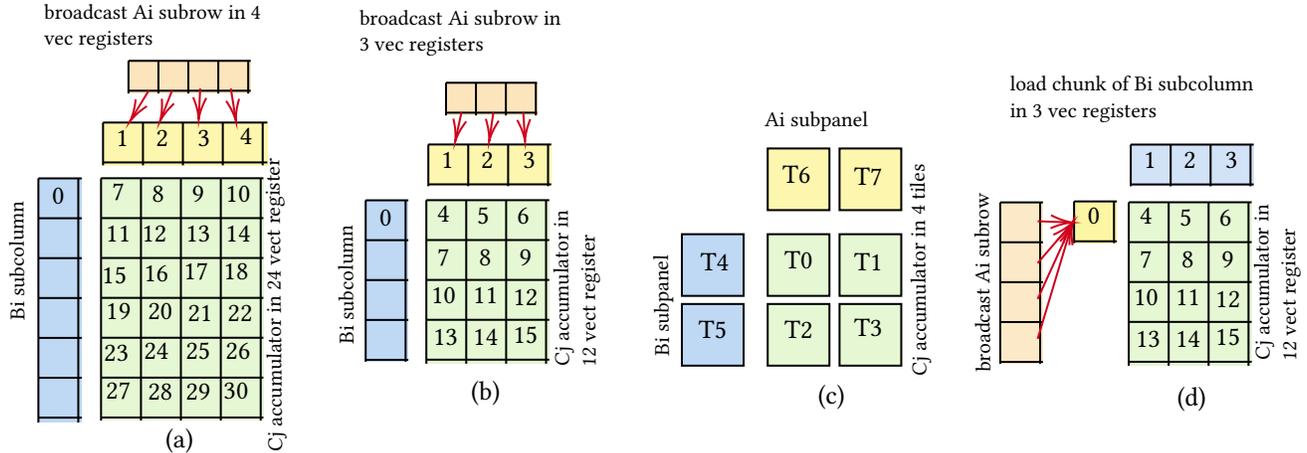} }
\vspace{-.20cm}
\caption{BRGEMM register tiling described in~\cite{tpp-tiling} for (a) 32 vector registers (b) 16 vector registers (c) 8 2D register tiles and (d) 16 vector registers with loads of A and B are swapped.}
\label{fig:resgister-block}
\vspace{-.20cm}
\end{figure*}

\subsection{BRGEMM Mixed Precision Operation With VNNI}

The rapid growth of AI and machine learning workloads has driven a shift from traditional FP32/FP64 formats toward mixed-precision types such as BF16~\cite{bf16}, FP16~\cite{f16}, and INT8, which improve computational efficiency and reduce memory footprint.
Among these, BF16 has gained prominence for its balanced trade-off between dynamic range and precision.
Achieving efficient BRGEMM execution with BF16, however, requires careful data layout design to fully leverage hardware acceleration features.

In x86 architectures, Intel’s AVX512 VNNI (Vector Neural Network Instructions) provides fused multiply-accumulate operations optimized for packed data formats.
Instructions such as {\tt TDPBF16PS}/{\tt VDPBF16PS} operate on tightly packed 16-bit inputs and accumulate results in 32-bit registers.
Standard row- or column-major layouts lead to non-contiguous memory accesses during dot-product computation, reducing cache locality and SIMD efficiency.
To mitigate this, input matrices are pre-packed—BF16 elements in pairs and INT8 elements in groups of four—to align with vector register widths (Figure~\ref{fig:vnni-packing}).
This packing enables efficient vector loads and avoids costly runtime shuffling; without it, data rearrangement overhead can significantly degrade performance.

In the following sections, we show how MLIR compiler infrastructure automates this packing process and generates nanokernels optimized for VNNI instructions.
By abstracting hardware-specific details through custom lowering strategies, MLIR enables scalable and portable mixed-precision kernel generation across x86 architectures.
Figure~\ref{fig:vnni-packing} illustrates how a flat 4$\times$4 row-major matrix is transformed into a 2$\times$4$\times$2 layout with a VNNI{=}2 dimension.
% Listing~\ref{code:input} presents MLIR code for a running example—a VNNI-packed batch-reduce GEMM of mixed precision (A/B matrices in BF16 and C matrix in FP32) with dimensions 32$\times$32$\times$192 (A$\times$C$\times$B).
%
Listing~\ref{code:input} presents a VNNI-packed BRGEMM represented using {\tt linalg.contract}. The {\tt affine\_map} at line~2 defines the dimension mapping for matrix~A, while line~3 and~4 define those for matrices~B and~C, respectively.
%
%with dimensions 32$\times$32$\times$192 (A$\times$C$\times$B).
The K dimension is packed with VNNI{=}2 for BF16 data.
Throughout the paper, we refer to this interleaved representation as the \textbf{VNNI-packed layout} and the conventional row/column-major order as the \textbf{Flat layout}.

\begin{lstlisting}[style=customC, basicstyle=\ttfamily\scriptsize, label=code:input, float, caption=Example BRGEMM with VNNI{=}2 packing. , belowskip=-16pt]
linalg.contract indexing_maps = [
 affine_map<(batch,m,n,k,vnni) -> (batch,m,k,vnni)>,
 affine_map<(batch,m,n,k,vnni) -> (batch,k,n,vnni)>,
 affine_map<(batch,m,n,k,vnni) -> (m, n)>]
 ins(%A, %B : memref<4x32x16x2xbf16>,
   memref<4x16x192x2xbf16>)
 outs(%C : memref<32x192xf32>)
\end{lstlisting}

\subsection{Microkernels and Nanokernels}
When writing high-performance kernels, developers often separate the code into two parts: (1) Memory tiling and iteration order; and (2) the inner compute loop, which we call the \textbf{microkernel}.
The inner-loop should have enough compute to fill an entire level of cache (L1/L2) with data (read and write), but that's more compute than can fit into registers, to keep the accumulation \textit{tile} in registers throughout the whole BRGEMM. So we further  tile the BRGEMM \textit{microkernel} into smaller BRGEMMs, introducing \textbf{nanokernels}.
The considerations for building the \textit{perfect} nanokernel involves deep knowledge of the micro-architecture (available extensions, number of registers, latency of instructions, cache associativity, multi-issue, out-of-order and pipelining).
Since we are using LLVM as our last-mile compiler, it also involves knowing how good the LLVM back-end is at recognizing certain patterns, to avoid spills and to allow software pipelining without having the backend rewrite our code.

\subsection{TPP-MLIR Compiler}
TPP-MLIR~\cite{tppmlir} (Tensor Processing Primitives – MLIR) is an AI compiler framework that generates high-performance CPU code for ML and scientific workloads.
Built on top of MLIR, it represents programs at multiple abstraction levels—from high-level tensor algebra to low-level machine code.
TPP-MLIR accepts models from frameworks like TensorFlow or PyTorch and converts them into a \textit{Linalg-on-Tensor} intermediate representation using the {\tt linalg} dialect.
It then applies hardware-agnostic optimizations, including tiling, fusion, and packing, through upstream MLIR transformation passes to improve data locality and parallelism.

For hardware-specific performance tuning, TPP-MLIR integrates with the \texttt{libxsmm} library, a highly optimized collection of kernels for tensor operations.
To enable this integration, the compiler lowers the optimized IR into its own {\tt XSMM} dialect, which serves as an interface to invoke the appropriate \texttt{libxsmm} kernels for the target architecture.
In the following sections, we present techniques for generating target-optimized kernels using MLIR dialects and demonstrate them with TPP-MLIR, without depending on \texttt{libxsmm} and removing the
need for {\tt XSMM} dialect.

\section{Register Tiling and Vectorization}
\label{sec:tile_vector}

We first apply register tiling to the BRGEMM so that computations fits entirely within vector registers, and then lower it to a {\tt vector.contract} operation.
Both tiling and vectorization rely solely on upstream MLIR transformation passes.

\subsection{Register Tiling}
We adopt the register-tiling strategy from Section~3.2.1 of~\cite{tpp-tiling}, applied across the M, N, K, and batch dimensions to %ensure each tiled kernel fits within available vector registers without 
ensure no register spills (with batch tiles always fixed to 1 and K with 1/VNNI/32).
Tile sizes are chosen per target architecture, considering register widths of 128–512 bits and 2D tiles up to 1~KB.
Figure~\ref{fig:resgister-block}(a) illustrates AVX512 tiling with 32 registers, each holding 16 FP32 elements.
Tile sizes M:4 (A subrow broadcasts across v1–v4) and N:64 (B subcolumn chunks via v0) yield 29 used vector registers—24 accumulators, 4 broadcasts, and 1 B load—without spills.
For AVX2 (Figure~\ref{fig:resgister-block}(b)), 16 registers hold 8 FP32 elements each.
Using M:3 and N:32 utilizes all registers: 12 accumulators + 3 broadcasts (A subrow) + 1 load (B subcoumn).

For Intel AMX (Figure~\ref{fig:resgister-block}(c)), eight 2D tiles store 16×16 FP32 or 16×32 BF16 elements.
Tiles T6/T7 and T4/T5 store A/B subpanels, while T0–T3 hold accumulations.
The optimal spill-free setup is M:32, N:32, K:32, reusing A/B tiles twice.

Register tiling depends on matrix and tile sizes.
For a $12\times24$ matrix on a 16-register machine, M:3, N:24 uses 13 registers efficiently; increasing to M:4, N:24 causes a spill (17 registers).
In such cases, we swap broadcast/load roles: loading B subcolumns into v1–v3, broadcasting A via v0, and accumulating into v4–v15 (Figure~\ref{fig:resgister-block}(d))—using all 16 registers with no spills.
% Listing~\ref{code:blocking} shows BRGEMM tiling with M:6, N:96, K:2 (VNNI), and batch-reduce:1, using {\tt linalg::tileLinalgOp}.
% The K (reduction) loop is innermost, followed by batch-reduce.
% Lines 1–4 define tiled loops; 5–7 create subviews of A, B, and C; and Line 8 performs sub-BRGEMM.

% \subsection{Vectorization}
% After tiling, we apply vectorization to lower the  BRGEMM to a contraction operation with \texttt{vector.transfer\_read} and \texttt{vector.transfer\_write} operations to load/store data from the subview.
% Listing~\ref{code:vectorization} shows the IR after applying the vectorization pass on top of the tiled BRGEMM (Listing~\ref{code:blocking}). The {\tt linalg.generic} is lowered to the {\tt vector.contract}, which serves as the starting point to our nanokernel generation.

\subsection{Vectorization}
Following register tiling, the BRGEMM operation is vectorized into a contraction form. Data movement is expressed through {\tt vector.transfer\_read/write}, while computation is carried out using {\tt vector.contract}, defining the entry point for our nanokernel generation.

\begin{lstlisting}[style=customC, basicstyle=\ttfamily\scriptsize, label=code:vectorization, float, caption=Input MLIR to our code generation after register tiling (with M:4 and N:96) and vectorization on Listing~\ref{code:input}., belowskip=-16pt]
scf.for %a3 = 0 to 32 step 4 // M-Loop
 scf.for %a4 = 0 to 192 step 96 // N-Loop
  %2=t_read %C[%a3,%a4], vector<4x96xf32> 
  %3 = scf.for %a5=0 to 4 step 1 args(%a7=%2){ //batch
   %4 = scf.for %a6=0 to 16 step 1 args(%a8=%a7){ //K
    %5=t_read %A[%a5,%a3,%a6,0], vector<1x4x1x2xbf16>
    %6=t_read %B[%a5,%a6,%a4,0], vector<1x1x96x2xbf16>
    %7=vector.contract ins(%5,%6:vector<1x4x1x2xbf16>, vector<1x1x96x2xbf16>)outs(%a8:vector<4x96xf32>)
    scf.yield : vector<4x96xf32> 
   }
   scf.yield : vector<4x96xf32>
  }
  t_write %3, %C[%a3,%a4], vector<4x96xf32>
\end{lstlisting}

\section{Nanokernels: Contraction Optimizations}
\label{sec:nano-kernels}

% In this section, we present a technique to lower vector contraction into target-specific nanokernels. 

% \subsection{MLIR Dialect Operations}

% We make use of two main upstream MLIR dialects in addition to the standard \texttt{vector} dialect, to lower our BRGEMM nanokernels: \texttt{amx} and \texttt{x86vector}.

In this section, we describe how vector contractions (shown in Listing~\ref{code:vectorization}) are lowered into target-specific nanokernels using carefully selected MLIR dialect operations.
Leveraging MLIR’s extensible dialect system, our approach expresses nanokernel computations through operations closely aligned with target instructions, ensuring efficient code generation and near hand-optimized performance across architectures.

\subsection{MLIR Dialect Operations}
Our lowering leverages three primary upstream MLIR dialects—\texttt{amx},  \texttt{x86vector}, and \texttt{vector} dialect. These dialects enable the generation of BRGEMM nanokernels specialized for vector based instruction sets.

 \subsubsection{AMX Operations} The {\tt AMX}~\cite{amx} dialect provides operations used for generating AMX-targeted nanokernels:
 \begin{itemize}
     \item {\tt amx.tile\_load/store} - load or store elements into 1 KB register tiles,
     \item {\tt amx.tile\_mulf} - perform tile dot product on \texttt{BF16} input tiles with accumulation in \texttt{FP32}.
 \end{itemize}

 %  %\begin{itemize}
 %     {\tt amx.tile\_load/store} - load or store elements into 1 KB register tiles
     
 %     {\tt amx.tile\_mulf} - perform tile dot product on \texttt{BF16} input tiles with accumulation in \texttt{FP32}.
 % %\end{itemize}

\subsubsection{AVX Packed Operations}  As part of this work, we upstreamed five operations into the \texttt{x86vector}~\cite{x86Vector} dialect. These operations map directly to LLVM intrinsics and ensure precise code generation for load, store, broadcast, and dot-product operations on packed data types:
 \begin{itemize}
     \item {\tt bcst\_to\_f32.packed} -  converts and broadcasts a scalar \texttt{BF16}/\texttt{F16} element into packed \texttt{FP32} elements,
     \item  {\tt even.indexed\_to\_f32} - converts the even-indexed \texttt{BF16}/\texttt{F16} elements into packed \texttt{FP32} elements,
     \item {\tt odd.indexed\_to\_f32} - converts the odd-indexed \texttt{BF16}/ \texttt{F16} elements into packed \texttt{FP32} elements.
     \item {\tt avx512.dot} - performs a dot-product on \texttt{BF16} inputs and accumulates results in \texttt{FP32}. 
    %  Syntax:\\
    % avx512.dot acc, a, b : vec$<$32$\times$bf16$>$$\rightarrow$vec$<$16$\times$f32$>$
     \item {\tt packed.f32\_to\_bf16} - converts packed \texttt{FP32} elements back to packed \texttt{BF16}. 
 \end{itemize}

\begin{figure}[t]
\centering
\resizebox{.40\textwidth}{!}{ \tikzset{every picture/.style={line width=0.75pt}} %set default line width to 0.75pt        

\begin{tikzpicture}[x=0.75pt,y=0.75pt,yscale=-1,xscale=1]
%uncomment if require: \path (0,168); %set diagram left start at 0, and has height of 168

%Shape: Grid [id:dp3148297696166734] 
\draw  [draw opacity=0][fill={rgb, 255:red, 184; green, 233; blue, 134 }  ,fill opacity=0.5 ] (34,22.19) -- (323.96,22.19) -- (323.96,42.08) -- (34,42.08) -- cycle ; \draw   (34,22.19) -- (34,42.08)(52,22.19) -- (52,42.08)(70,22.19) -- (70,42.08)(88,22.19) -- (88,42.08)(106,22.19) -- (106,42.08)(124,22.19) -- (124,42.08)(142,22.19) -- (142,42.08)(160,22.19) -- (160,42.08)(178,22.19) -- (178,42.08)(196,22.19) -- (196,42.08)(214,22.19) -- (214,42.08)(232,22.19) -- (232,42.08)(250,22.19) -- (250,42.08)(268,22.19) -- (268,42.08)(286,22.19) -- (286,42.08)(304,22.19) -- (304,42.08)(322,22.19) -- (322,42.08) ; \draw   (34,22.19) -- (323.96,22.19)(34,40.19) -- (323.96,40.19) ; \draw    ;
%Shape: Grid [id:dp3505796703763506] 
\draw  [draw opacity=0][fill={rgb, 255:red, 74; green, 144; blue, 226 }  ,fill opacity=0.25 ] (158,61.19) -- (324,61.19) -- (324,78.6) -- (158,78.6) -- cycle ; \draw   (158,61.19) -- (158,78.6)(173,61.19) -- (173,78.6)(188,61.19) -- (188,78.6)(203,61.19) -- (203,78.6)(218,61.19) -- (218,78.6)(233,61.19) -- (233,78.6)(248,61.19) -- (248,78.6)(263,61.19) -- (263,78.6)(278,61.19) -- (278,78.6)(293,61.19) -- (293,78.6)(308,61.19) -- (308,78.6)(323,61.19) -- (323,78.6) ; \draw   (158,61.19) -- (324,61.19)(158,76.19) -- (324,76.19) ; \draw    ;
%Shape: Grid [id:dp4076787851767293] 
\draw  [draw opacity=0][fill={rgb, 255:red, 189; green, 16; blue, 224 }  ,fill opacity=0.25 ] (124,101.19) -- (270,101.19) -- (270,119.6) -- (124,119.6) -- cycle ; \draw  [color={rgb, 255:red, 0; green, 0; blue, 0 }  ,draw opacity=1 ] (124,101.19) -- (124,119.6)(142,101.19) -- (142,119.6)(160,101.19) -- (160,119.6)(178,101.19) -- (178,119.6)(196,101.19) -- (196,119.6)(214,101.19) -- (214,119.6)(232,101.19) -- (232,119.6)(250,101.19) -- (250,119.6)(268,101.19) -- (268,119.6) ; \draw  [color={rgb, 255:red, 0; green, 0; blue, 0 }  ,draw opacity=1 ] (124,101.19) -- (270,101.19)(124,119.19) -- (270,119.19) ; \draw  [color={rgb, 255:red, 0; green, 0; blue, 0 }  ,draw opacity=1 ]  ;
%Shape: Grid [id:dp4608037563810211] 
\draw  [draw opacity=0][fill={rgb, 255:red, 245; green, 166; blue, 35 }  ,fill opacity=0.4 ] (65,143.19) -- (210,143.19) -- (210,161.6) -- (65,161.6) -- cycle ; \draw   (65,143.19) -- (65,161.6)(83,143.19) -- (83,161.6)(101,143.19) -- (101,161.6)(119,143.19) -- (119,161.6)(137,143.19) -- (137,161.6)(155,143.19) -- (155,161.6)(173,143.19) -- (173,161.6)(191,143.19) -- (191,161.6)(209,143.19) -- (209,161.6) ; \draw   (65,143.19) -- (210,143.19)(65,161.19) -- (210,161.19) ; \draw    ;
%Straight Lines [id:da9949829968277385] 
\draw    (44,39.6) -- (156.07,69.09) ;
\draw [shift={(158,69.6)}, rotate = 194.74] [color={rgb, 255:red, 0; green, 0; blue, 0 }  ][line width=0.75]    (10.93,-3.29) .. controls (6.95,-1.4) and (3.31,-0.3) .. (0,0) .. controls (3.31,0.3) and (6.95,1.4) .. (10.93,3.29)   ;
%Straight Lines [id:da040635502735341555] 
\draw    (44,39.6) -- (122.5,109.27) ;
\draw [shift={(124,110.6)}, rotate = 221.59] [color={rgb, 255:red, 0; green, 0; blue, 0 }  ][line width=0.75]    (10.93,-3.29) .. controls (6.95,-1.4) and (3.31,-0.3) .. (0,0) .. controls (3.31,0.3) and (6.95,1.4) .. (10.93,3.29)   ;
%Straight Lines [id:da3039960200117745] 
\draw    (44,39.6) -- (64.6,141.23) ;
\draw [shift={(65,143.19)}, rotate = 258.54] [color={rgb, 255:red, 0; green, 0; blue, 0 }  ][line width=0.75]    (10.93,-3.29) .. controls (6.95,-1.4) and (3.31,-0.3) .. (0,0) .. controls (3.31,0.3) and (6.95,1.4) .. (10.93,3.29)   ;

% Text Node
\draw (54,23.19) node [anchor=north west][inner sep=0.75pt]   [align=left] {1};
% Text Node
\draw (160,61.19) node [anchor=north west][inner sep=0.75pt]   [align=left] {0};
% Text Node
\draw (176,61) node [anchor=north west][inner sep=0.75pt]   [align=left] {0};
% Text Node
\draw (191,61) node [anchor=north west][inner sep=0.75pt]   [align=left] {0};
% Text Node
\draw (206,61) node [anchor=north west][inner sep=0.75pt]   [align=left] {0};
% Text Node
\draw (220,61) node [anchor=north west][inner sep=0.75pt]   [align=left] {0};
% Text Node
\draw (236,61) node [anchor=north west][inner sep=0.75pt]   [align=left] {0};
% Text Node
\draw (251,61) node [anchor=north west][inner sep=0.75pt]   [align=left] {0};
% Text Node
\draw (266,61) node [anchor=north west][inner sep=0.75pt]   [align=left] {0};
% Text Node
\draw (281,61) node [anchor=north west][inner sep=0.75pt]   [align=left] {0};
% Text Node
\draw (296,61) node [anchor=north west][inner sep=0.75pt]   [align=left] {0};
% Text Node
\draw (310,62.19) node [anchor=north west][inner sep=0.75pt]   [align=left] {0};
% Text Node
\draw (39,23) node [anchor=north west][inner sep=0.75pt]   [align=left] {0};
% Text Node
\draw (126,103) node [anchor=north west][inner sep=0.75pt]   [align=left] {0};
% Text Node
\draw (67,143.19) node [anchor=north west][inner sep=0.75pt]   [align=left] {1};
% Text Node
\draw (93,23.19) node [anchor=north west][inner sep=0.75pt]   [align=left] {3};
% Text Node
\draw (74,23) node [anchor=north west][inner sep=0.75pt]   [align=left] {2};
% Text Node
\draw (127,24.19) node [anchor=north west][inner sep=0.75pt]   [align=left] {5};
% Text Node
\draw (110,23) node [anchor=north west][inner sep=0.75pt]   [align=left] {4};
% Text Node
\draw (163,23.19) node [anchor=north west][inner sep=0.75pt]   [align=left] {7};
% Text Node
\draw (146,24) node [anchor=north west][inner sep=0.75pt]   [align=left] {6};
% Text Node
\draw (181,23.19) node [anchor=north west][inner sep=0.75pt]   [align=left] {8};
% Text Node
\draw (212,23.19) node [anchor=north west][inner sep=0.75pt]   [align=left] {10};
% Text Node
\draw (199,23) node [anchor=north west][inner sep=0.75pt]   [align=left] {9};
% Text Node
\draw (231,23.19) node [anchor=north west][inner sep=0.75pt]   [align=left] {11};
% Text Node
\draw (247,23.19) node [anchor=north west][inner sep=0.75pt]   [align=left] {12};
% Text Node
\draw (266,23.19) node [anchor=north west][inner sep=0.75pt]   [align=left] {13};
% Text Node
\draw (284,23.19) node [anchor=north west][inner sep=0.75pt]   [align=left] {14};
% Text Node
\draw (302,23.19) node [anchor=north west][inner sep=0.75pt]   [align=left] {15};
% Text Node
\draw (146,102) node [anchor=north west][inner sep=0.75pt]   [align=left] {2};
% Text Node
\draw (164,102) node [anchor=north west][inner sep=0.75pt]   [align=left] {4};
% Text Node
\draw (182,102) node [anchor=north west][inner sep=0.75pt]   [align=left] {6};
% Text Node
\draw (200,102.19) node [anchor=north west][inner sep=0.75pt]   [align=left] {8};
% Text Node
\draw (213,102.19) node [anchor=north west][inner sep=0.75pt]   [align=left] {10};
% Text Node
\draw (230,102.19) node [anchor=north west][inner sep=0.75pt]   [align=left] {12};
% Text Node
\draw (249,103.19) node [anchor=north west][inner sep=0.75pt]   [align=left] {14};
% Text Node
\draw (87,144.19) node [anchor=north west][inner sep=0.75pt]   [align=left] {3};
% Text Node
\draw (105,144.19) node [anchor=north west][inner sep=0.75pt]   [align=left] {5};
% Text Node
\draw (124,144.19) node [anchor=north west][inner sep=0.75pt]   [align=left] {7};
% Text Node
\draw (142,144) node [anchor=north west][inner sep=0.75pt]   [align=left] {9};
% Text Node
\draw (156,143.19) node [anchor=north west][inner sep=0.75pt]   [align=left] {11};
% Text Node
\draw (173,144.19) node [anchor=north west][inner sep=0.75pt]   [align=left] {13};
% Text Node
\draw (190,144.19) node [anchor=north west][inner sep=0.75pt]   [align=left] {15};
% Text Node
\draw (35,6) node [anchor=north west][inner sep=0.75pt]  [font=\normalsize] [align=left] {memref$<$16xbf16$>$};
% Text Node
\draw (230,47) node [anchor=north west][inner sep=0.75pt]  [font=\normalsize] [align=left] {vector$<$8xf32$>$};
% Text Node
\draw (176,87) node [anchor=north west][inner sep=0.75pt]  [font=\normalsize] [align=left] {vector$<$8xf32$>$};
% Text Node
\draw (114,128) node [anchor=north west][inner sep=0.75pt]  [font=\normalsize] [align=left] {vector$<$8xf32$>$};
% Text Node
\draw (76,51) node [anchor=north west][inner sep=0.75pt]  [font=\small] [align=left] {bcst\_to\_f32};
% Text Node
\draw (61,83) node [anchor=north west][inner sep=0.75pt]  [font=\small] [align=left] {even.indx\_to\_f32};
% Text Node
\draw (16,115) node [anchor=north west][inner sep=0.75pt]  [font=\small] [align=left] {odd.indx\_to\_f32};

\end{tikzpicture} }
\caption{AVX2 BF16 Packed  Operations}
\label{fig:pack-ins}
\vspace{-.20cm}
\end{figure}
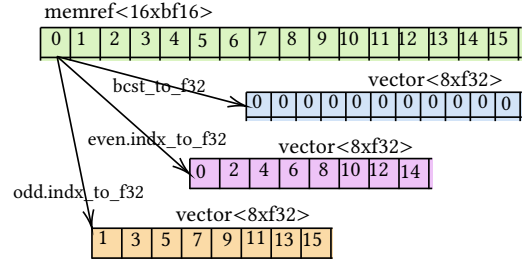

Figure~\ref{fig:pack-ins} illustrates how the $0th$ element of a \texttt{memref} is converted and broadcast to a \texttt{vector<8xf32>} using \texttt{bcst\_to\_f32.packed}. Similarly, the even and odd indexed elements of the same \texttt{memref} are converted and packed into a \texttt{vector<8xf32>}  using the \texttt{even/odd.indexed\_to\_f32} operations.
% We also rely on generic operations from the \texttt{vector} and \texttt{arith} dialects, including \texttt{vector.load/store},   \texttt{vector.fma}, \texttt{vector.shuffle}, \texttt{vector.bitcast}, 
% \texttt{arith.shli}, \texttt{arith.andi}, and \texttt{vector.broadcast}.
%
We also rely on generic operations from the \texttt{vector} and \texttt{arith} dialects, such as \texttt{vector.load/store}, \texttt{vector.fma}, \texttt{vector.shuffle}, \texttt{arith.shli}, and others.

\begin{table}
\centering
\begin{tabular}{l |l |l |l} 
 \hline
 \textbf{Type} & \textbf{Target} & \textbf{Matrix Data Op} & \textbf{Matmul Op} \\
 \hline
 \multirow{8}{*}{BF16} 
 & AMX & {\tt amx.tile\_load} & {\tt amx.tile\_mulf} \\
 & & {\tt amx.tile\_store} &  \\
\cline{2-4}
 & AVX512 & {\tt vector.load} & {\tt avx512.dot}{*} \\
 & \_BF16 & {\tt vector.store} &  \\
 \cline{2-4}
   & AVX2$^{+}$ & {\tt even.indx\_to\_f32}{*} & {\tt vector.fma} \\
 &  & {\tt odd.indx\_to\_f32}{*} &  \\
 \cline{2-4}
  & any ISA$^{+}$  & {\tt vector.load/store} & {\tt vector.fma} \\
  &  & {\tt bitcast,shli,andi} &  \\
\hline
\multirow{2}{*}{FP32}
 & any ISA & {\tt vector.load} & {\tt vector.fma} \\
 & & {\tt vector.store} &  \\
 \hline
\end{tabular}
\caption{Mapping of MLIR dialect operations to target machines used in our nanokernel code generation process. {*} - {\tt x86vector} dialect operations; $^{+}$ - Emulation lowering.}
\label{table:mlir-op}
\vspace{-0.9cm}
\end{table}

Table~\ref{table:mlir-op} shows the most suitable mapping between MLIR dialect operations and target architectures used in our code generation process: AMX uses tile registers for block loads and dot-product; AVX512 leverages vector load/store with the {\tt avx512.dot} intrinsic; AVX2 uses indexed conversion with {\tt vector.fma} for emulation; and generic targets without BF16 support rely on type-conversion primitives combined with {\tt vector.fma}.
For FP32 workloads, the same vector operations are reused across all ISAs, ensuring portability and consistency in code generation.

\subsection{Generalized View of Nanokernel Generation}
Algorithm~\ref{alg:generic} illustrates the high-level structure of nanokernel generation based on the data type and target machine.
The C subblock is loaded (line~3) before the innermost tiled loops (batch-reduce and K), where we iterate over these loops, accumulate the multiplied results of A and B, and store the result back to the C subblock after completion (line~19).
For FP32 nanokernels, the generation process isindependent of the target machine, as shown in lines~8–12.
All elements of the A subrow $A_{i=0..M}$ are loaded and broadcast into M vector registers, followed by iterating over the chunks of the B subcolumn, loading each chunk, and computing fused multiply–add (FMA) operations using {\tt vector.fma}\footnote{Due to space constraints, we present the generated {\tt FP32} and {\tt BF16} nanokernel IR in the appendix.}.

For BF16 nanokernels, the code generation process differs depending on the input matrix layout and target hardware.
As discussed in Section~\ref{sec:background}, the input matrix must be VNNI-packed for the dot-product operation to efficiently load and compute.
If the input remains in a flat layout (row- or column-major), it must be packed into VNNI format before computation.
We implement this transformation efficiently using {\tt vpunpcklwd} and {\tt vpunpckhwd} instructions, which perform low-latency 16-bit shuffles ideally suited for BF16 data.

\SetKw{KwStep}{step}
\begin{algorithm}
\caption{BRGEMM - nanokernels()}
\label{alg:generic}
\KwIn{$A^{M \times K}_i, B^{K \times N}_i$ for $i = 0, .. n-1, C^{M \times N}, \beta \in \mathbb{R}$ }
\KwOut{$C = \beta . C + \sum_{i=0}^{n-1} A_i \times B_i$}
\For{$i_m \gets 0$ \KwTo $m-1$ \KwStep $mb$}{
  \For{$i_n \gets 0$ \KwTo $n-1$ \KwStep $nb$}{
    $acc_{m \times n}$ = load $m_{b} \times n_{b}$ C-subblock$_{im, in}$\;
    \If{$Flat$ and target is $AMX$}{
            packB = S/w-pipeline + pack using $vpunpck$ the $0^{th}$ $i_{br}$ $B_{in}$ subpanel\;
    }
    \For{$i_{br} \gets 0$ \KwTo $BR-1$ \KwStep $1$}{
      \For{$i_{k} \gets 0$ \KwTo $k-1$ \KwStep $vnni$}{

        \If{$type$ is $f32$}{
            %GEMM-nano kernels f32()\;
            bcstA$_{i=0..m}$ = bcast $A_{i = 0..m}$ in $M$\;% vec registers\;
            \For{$i_n \gets 0$ \KwTo $nb$ \KwStep $chunk$}{
                load $B_{in}$ in a vec register\;
                $acc_{im \times in} = vector.fma$ bcstA$_{i=0..m}$, $B_{in}$,  $acc_{im \times in}$\;
            }
        }
        
        \If{$type$ is $bf16$ and $VNNI$}{
            nanokernels VNNI bf16()\;
        }
    
        \If{$type$ is $bf16$ and $Flat$}{
            nanokernels Flat bf16()\;
        }
      }
    }
    \If{$Flat$} {
        Shuffle the final accumulator $acc_{m \times n}$\;
    }
    
    C-subblock$_{im, in}$ store $acc_{m \times n}$\;
  } 
}
\end{algorithm}

\begin{figure}%[t]
\centering
\resizebox{.5\textwidth}{!}{ \input{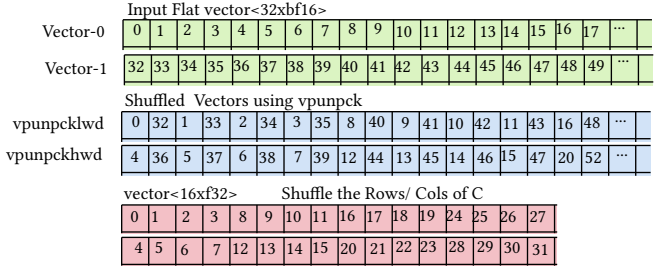} }
\vspace{-.75cm}
\caption{Converting Flat to VNNI packed layout using {\tt vpunpcklwd/hwd} operations.}
\label{fig:shuffle}
\vspace{-.40cm}
\end{figure}

Figure~\ref{fig:shuffle} shows the conversion from a flat to a VNNI packed layout for two {\tt vector<32×bf16>} registers using {\tt vpunpck}.
Each element in the original flat vectors (green) is rearranged into the packed format (blue).
A side effect of this approach is the need to shuffle the final accumulators (pink) to match the interleaved structure (lines~17–18 in Algorithm~\ref{alg:generic}).
% In the following sub-sections, we detail BF16 nanokernel generation for both VNNI-packed and flat input formats.

\begin{algorithm}
\caption{BRGEMM nanokernels VNNI bf16()}
\label{alg:bf16+vnni}
\KwIn{vector.contract $\xrightarrow{}$ $A^{M \times K}_i, B^{K \times N}_i$}
\KwOut{nanokernels with $M \times N$ FMAs or DPs}
   \If{$target$ is $AMX$}{
     tileA = $amx.tile\_load$  $A_{i=0..m}$\;
     tileB = $amx.tile\_load$  $B_{i=0..n}$\;
     $amx.tile\_mulf$ tileA$_{i=0..m}$, tileB$_{i=0..n}$, $acc_{im \times in}$\;
   }
   \ElseIf{$target$ has $bf16dp$}{
     bcstA$_{i=0..m}$ =  bcast $A_{i=0..m}$ in $M$ vec registers\;
     \For{$i_n \gets 0$ \KwTo $nb$ \KwStep $chunk$}{
        load $B_{in}$ in a vec register\;
        $acc_{im \times in}$ = $avx512.dot$ bcstA$_{i=0..m}$, $B_{in}$,  $acc_{im \times in}$\;
    }
   }
   \ElseIf{$target$ has AVX2 even and odd pack ins}{
     bcstA$_{i=0..m}$ = bcast\_odd\_to\_f32 $A_{i=0..m}$ in $M$\;
    \For{$i_n \gets 0$ \KwTo $nb$ \KwStep $chunk$}{
        odd\_load $B_{in}$ in a vec register\;
        $odd\_acc_{im \times in} = vector.fma$ bcstA$_{i=0..m}$, $oddB_{in}$,  $acc_{im \times in}$\;
    }
    bcstA$_{i=0..m}$ = bcast\_even\_to\_f32 $A_{i=0..m}$ in $M$\;
    \For{$i_n \gets 0$ \KwTo $nb$ \KwStep $chunk$}{
        even\_load $B_{in}$ in a vec register\;
        $acc_{im \times in} = vector.fma$ bcstA$_{i=0..m}$, $evenB_{in}$, $odd\_acc_{im \times in}$\;
    }
   }
   \Else{
   load $A_{i=0..m}$, bitcast and bcast it in $M$ registers\;
   perform $andi$ to get the odd elements of $A_{i=0..m}$\;
   \For{$i_n \gets 0$ \KwTo $nb$ \KwStep $chunk$}{
        load $B_{in}$ in a vec register\;
        perform $andi$ to get the odd elements\;
        $odd\_acc_{im \times in} = vector.fma$ bcstA$_{i=0..m}$, $oddB_{in}$,  $acc_{im \times in}$\;
    }

  load $A_{i=0..m}$, bitcast and bcast it in $M$ registers\;
   perform $shli$ to get the even elements of $A_{i=0..m}$\;
   \For{$i_n \gets 0$ \KwTo $nb$ \KwStep $chunk$}{
        load $B_{in}$ in a vec register\;
        perform $shiftl$ to get the even elements\;
        $acc_{im \times in} = vector.fma$ bcstA$_{i=0..m}$, $evenB_{in}$, $odd\_acc_{im \times in}$\;
    }
   }

\end{algorithm}
%\vspace{-0.9cm}

\begin{algorithm}
\caption{BRGEMM nanokernels Flat bf16()}
\label{alg:bf16+splat}
\KwIn{vector.contract $\xrightarrow{}$ $A^{M \times K}_i, B^{K \times N}_i$}
\KwOut{nanokernels with $M \times N$ FMAs or DPs}
   \If{$target$ is $AMX$}{
     tileA = $amx.tile\_load$  $A_{i=0..m}$\;
     tileB = $amx.tile\_load$  packB$_{i}$\;
     $amx.tile\_mulf$ tileA$_{i=0..m}$, tileB$_{i=0..n}$, $acc_{im \times in}$\;
     packB$_{i}$ = Sw-pipeline + pack using $vpunpck$ the $i_{br+1}$ or $i_{k+1}$ B subpanel\;
   }
   \ElseIf{$target$ has $bf16dp$}{
     bcstA$_{i=0..m}$ =  bcast $A_{i=0..m}$ in $M$ vec registers\;
     \For{$i_n \gets 0$ \KwTo $nb$ \KwStep $2$$\times$$chunk$}{
        load $B_{in}$ and $B_{in + 1}$ in two vec register\;
        Pack $B_{in}$ and $B_{in + 1}$ using $vpunpck$\;
        $acc_{im \times in}$ = $avx512.dot$ bcstA$_{i=0..m}$, pack$B_{in}$,  $acc_{im \times in}$\;
        $acc_{im+1 \times in+1}$ = $avx512.dot$ bcstA$_{i=0..m}$, pack$B_{in+1}$,  $acc_{im+1 \times in+1}$\;
    }
   }
   \ElseIf{$target$ has AVX2 even and odd pack ins}{
     bcstA$_{i=0..m}$ = bcast\_to\_f32 $A_{i=0..m}$ in $M$\;
    \For{$i_n \gets 0$ \KwTo $nb$ \KwStep $chunk$}{
        even\_load $B_{in}$ in a vec register\;
        $acc_{im \times in}$ = $vector.fma$ bcstA$_{i=0..m}$, $evenB_{in}$,  $acc_{im \times in}$\;
        odd\_load $B_{in}$ in a vec register\;
        $acc_{im+1 \times in+1}$ = $vector.fma$ bcstA$_{i=0..m}$, $oddB_{in+1}$,  $acc_{im+1 \times in+1}$\;
    }
   }
   \Else{
   load $A_{i=0..m}$, bitcast and bcast it in $M$ registers\;
   \For{$i_n \gets 0$ \KwTo $nb$ \KwStep $chunk$}{
        load $B_{in}$ in a vec register\;
        perform $shli$ to get the even elements\;
        $acc_{im \times in}$ = $vector.fma$ bcstA$_{i=0..m}$, $evenB_{in}$,  $acc_{im \times in}$\;
        perform $andi$ to get the odd elements\;
        $acc_{im+1 \times in+1}$ = $vector.fma$ bcstA$_{i=0..m}$, $oddB_{in+1}$,  $acc_{im+1 \times in+1}$\;
    }
   }
%\vspace{-0.2cm}
\end{algorithm}

\subsection{BF16 AMX Nanokernel Generation}
\subsubsection{VNNI-Packed}
On Intel AMX targets, we load 16 rows or columns of the A and B subpanels using the {\tt amx.tile\_load} operation, followed by tiled dot-product computation with {\tt amx.tile\_multf} (lines~1–4 in Algorithm~\ref{alg:bf16+vnni}).
The accumulated results are written back to the C subblock using {\tt amx.tile\_store}.
If the C subblock is in BF16 format, it must first be up-converted to FP32 for accumulation and then down-converted back to BF16 before storing.
Listing~\ref{code:amx-vnni} shows the AMX-generated VNNI nanokernel assembly for the blocking configuration in Figure~\ref{fig:resgister-block}(c).
Lines~1–4 load 16 rows of 32$\times$bf16 elements (A subrows and B subcolumns) into registers $\%tmm4$–$\%tmm7$, while lines~5–8 perform the tiled dot-product and accumulate the results in $\%tmm0$–$\%tmm3$.

\paragraph{Flat Layout.}
As discussed, the input must be VNNI-packed for dot-product computation.
Depending on matrix order, one input is packed—B for row-major layouts and A for column-major layouts.
Assuming row-major order, the A subpanel load remains identical to the VNNI-packed case, while the B subpanel is packed using {\tt vpunpck} instructions.
With a register tile size of 1~KB, we exploit software pipelining by overlapping packing of the B subpanel with the {\tt amx.tile\_multf} computation.
The first software-pipelined packing of the $0^{th}$ batch-reduce iteration is performed before the tiled loop (Algorithm~\ref{alg:generic}, lines~4–5).
During nanokernel execution, the packed B subpanel is loaded, the dot-product is computed, and the next B subpanel simultaneously packed in the background (Algorithm~\ref{alg:bf16+splat}, lines~2–5).
Listing~\ref{code:amx-splat} shows the AMX-generated flat nanokernel assembly: lines~1–5 loop over 16 rows and pack two {\tt vector<32×bf16>} chunks of the B subpanel using {\tt vpunpcklwd} and {\tt vpunpckhwd}; lines~12–13 load the packed B subpanel into tile registers.
The remaining instructions are identical to the VNNI-packed case.

\subsection{BF16 AVX Nanokernel Generation}
\subsubsection{DotBF16Op Nanokernels}
If the target machine supports {\tt VDPBF16PS}, the nanokernel generation process is similar to FP32.
Each pair (2$\times$bf16 elements) of the A$_{i=0..M}$ subrow is loaded and broadcast into M vector registers.
We then iterate over chunks of the B subcolumn, load each chunk, compute the dot-product using {\tt x86vector.avx512.dot}, and accumulate in FP32 (lines~6–9 in Algorithm~\ref{alg:bf16+vnni}).
Listing~\ref{code:bf16dp-vnni} shows the assembly of DotBF16Op-generated VNNI-packed nanokernels for the blocking configuration in Figure~\ref{fig:resgister-block}(a).
Lines~1, 4, 6, and 8 broadcast element pairs of the A subrow into registers $\%zmm24$ and $\%zmm26$–$\%zmm28$.
Line~2 loads the first B subcolumn chunk into $\%zmm25$, computes four dot-products (lines~3, 5, 7, and 9), and accumulates results in $\%zmm0$–$\%zmm3$.
Lines~16 and above repeat the process for the next chunk, with accumulations stored across $\%zmm20$–$\%zmm23$.

\begin{lstlisting}[style=customC, basicstyle=\ttfamily\scriptsize, label=code:amx-vnni, float, caption=BF16 AMX VNNI packed nanokernels. , belowskip=-12pt]
tileloadd (%r15,%r10), %tmm4 // 0th A's amx.tile_load 
tileloadd (%r13,%r10), %tmm5 // 1st A's amx.tile_load
tileloadd (%r11,%rdi), %tmm6 // 0th B's amx.tile_load
tileloadd (%r13,%rdi), %tmm7 // 1st B's amx.tile_load
tdpbf16ps %tmm6, %tmm4, %tmm0 // 0th amx.multf
tdpbf16ps %tmm7, %tmm4, %tmm1 // 1st amx.multf
tdpbf16ps %tmm6, %tmm5, %tmm2 // 2nd amx.multf
tdpbf16ps %tmm7, %tmm5, %tmm3 // 3rd amx.multf
... // assembly skipped for space
tilestored %tmm0,(%rax,%rdi) //0th acc amx.tile_store 
\end{lstlisting}

\begin{lstlisting}[style=customC, basicstyle=\ttfamily\scriptsize, label=code:amx-splat, float, caption=BF16 AMX Flat nanokernels., belowskip=-12pt]
.LBB0_2:
vmovdqu64 (%rcx,%rdx,2), %zmm0 //Load ith chunk of B
vmovdqu64 64(%rcx,%rdx,2),%zmm1 //Load i+1 chunk of B
vpunpcklwd %zmm1, %zmm0, %zmm2 // shuffle lower bits of ith and i+1th chunk of B
vpunpckhwd %zmm1, %zmm0, %zmm0  // shuffle lower bits of ith and i+1th chunk of B
... // assembly skipped for space

.LBB0_4:
jg .LBB0_2
tileloadd (%r14,%r11), %tmm5
tileloadd (%rbx,%r11), %tmm4
tileloadd (%r14,%r12), %tmm6 // Load from the pack-0
tileloadd (%rbx1,%r12), %tmm7 // Load from the pack-1
tdpbf16ps %tmm6, %tmm5, %tmm0
... // assembly skipped for space
\end{lstlisting}

\begin{lstlisting}[style=customC, basicstyle=\ttfamily\scriptsize, label=code:bf16dp-vnni, float, caption=DotBF16Op VNNI packed nanokernels., belowskip=-12pt]
vbroadcastss (%rsi,%rdx,4), %zmm24 // A[0] broadcast
vmovups -320(%rcx), %zmm25 // Load 0th chunk of B
vdpbf16ps %zmm25, %zmm24, %zmm1 // 0th avx512.dot
vbroadcastss 64(%rsi,%rdx,4), %zmm26 // A[1] broadcast
vdpbf16ps %zmm25, %zmm26, %zmm0 // 1st avx512.dot
vbroadcastss 128(%rsi,%rdx,4), %zmm27 //A[2] broadcast
vdpbf16ps %zmm25, %zmm27, %zmm3 // 2nd avx512.dot
vbroadcastss 192(%rsi,%rdx,4), %zmm28 //A[3] broadcast
vdpbf16ps %zmm25, %zmm28, %zmm2 // 3rd avx512.dot
vmovups -256(%rcx), %zmm25 // Load 1st chunk of B
vdpbf16ps %zmm25, %zmm24, %zmm5 // 4th avx512.dot
vdpbf16ps %zmm25, %zmm26, %zmm4 // 5th avx512.dot
vdpbf16ps %zmm25, %zmm27, %zmm7 // 6th avx512.dot
vdpbf16ps %zmm25, %zmm28, %zmm6 // 7th avx512.dot
.... // assembly skipped for space
vmovups (%rcx), %zmm25 // Load 5th chunk of B
vdpbf16ps %zmm25, %zmm24, %zmm21 // 20th avx512.dot
\end{lstlisting}

\paragraph{Flat Layout.}
For flat layouts, the B matrix must first be packed into VNNI format before computing dot-products using {\tt x86vector.avx512.dot} (Algorithm~\ref{alg:bf16+splat}, lines~7–12).
Since the B subcolumn is processed in pairs of chunks, one extra vector register is required.
For tile size M:4, N:64, register usage is 24 for accumulators, 4 for A broadcasts, and 2 for loading and packing consecutive B chunks—30 registers in total.
Listing~\ref{code:bf16dp-splat} shows the DotBF16Op generated Flat nanokernel assembly.
Lines~4–8 load two {\tt 32×bf16} chunks of the B subcolumn and pack them using {\tt vpunpcklwd} and {\tt vpunpckhwd}.
The packed data in $\%zmm30$ and $\%zmm28$ is used for dot-products against broadcasted A values.
The remaining instructions are identical to the VNNI-packed version.
Finally, the accumulated dot-products in $\%zmm20$–$\%zmm23$ are shuffled using {\tt vector.shuffle} to match the packed layout before being written back to the C subblock (lines~14–15 in Algorithm~\ref{alg:generic}).
Lines~25–28 in Listing~\ref{code:bf16dp-splat} perform this shuffle (0th and 4th dot-products) and store the results into the C matrix.

\begin{lstlisting}[style=customC, basicstyle=\ttfamily\scriptsize, label=code:bf16dp-splat, float, caption=DotBF16Op Flat nanokernels., belowskip=-14pt]
vbroadcastss (%rsi,%rdx,2), %zmm24 // A[0] broadcast
... // assembly skipped for space
vbroadcastss 192(%rsi,%rdx,2),%zmm27 //A[3] broadcast
vmovdqu64 -320(%rcx), %zmm28 // Load 0th chunk of B
vmovdqu64 -128(%rcx), %zmm29 // Load 1st chunk of B
vmovdqu64 (%rcx), %zmm31
vpunpcklwd %zmm29, %zmm28, %zmm30  // shuffle lower bits between 0th and 1st chunk of B
vpunpckhwd %zmm29,%zmm28,%zmm28  // shuffle higher bits between 0th and 1st chunk of B
vdpbf16ps %zmm28, %zmm24, %zmm23 // 0th avx512.dot
... // assembly skipped for space
vdpbf16ps %zmm30, %zmm24, %zmm21 // 4th avx512.dot
... // assembly skipped for space
vmovdqu64 -256(%rcx), %zmm28 // Load 4th chunk of B
vmovdqu64 -192(%rcx), %zmm29 // Load 5th chunk of B
vpunpcklwd %zmm31, %zmm29, %zmm30 
vpunpckhwd %zmm31, %zmm29, %zmm28 
vdpbf16ps %zmm30, %zmm24, %zmm18 // 16th avx512.dot
... // assembly skipped for space
vdpbf16ps %zmm28, %zmm24, %zmm19// 20th avx512.dot
vdpbf16ps %zmm28, %zmm25, %zmm13
... // assembly skipped for space
vmovapd .LCPI0_0(%rip), %zmm24
vmovapd .LCPI0_1(%rip), %zmm25
vmovaps %zmm21, %zmm26
vpermt2pd %zmm23,%zmm24,%zmm26//shuffle the 0, 4 acc
vpermt2pd %zmm23,%zmm25,%zmm21//shuffle the 0, 4 acc
vmovups %zmm26, (%rax) // store
vmovups %zmm21, 64(%rax) //store
... // assembly skipped for space
\end{lstlisting}

\subsubsection{AVX2 Packed-Operations Nanokernels}
When the target machine lacks native BF16 dot-product support but provides AVX2 packed operations for loading, storing, and broadcasting BF16 data, we emulate the nanokernel using FP32 computation.
Input BF16 elements are up-converted to FP32, and fused multiply–add (FMA) operations are performed using {\tt vector.fma}.

Two rounds of FMAs, termed \textit{odd} and \textit{even}, are executed.
For the \textit{odd} phase, the odd-indexed elements of each A$_{i=0..M}$ subrow (BF16) are loaded and broadcast into {\tt vector<8×f32>} using {\tt bcst\_to\_f32.packed}.
Chunks of the B subcolumn are then processed by loading their odd-indexed elements with {\tt odd.indexed\_to\_f32} followed by FMA (lines~11–14 in Algorithm~\ref{alg:bf16+vnni}).
For the \textit{even} phase, the even-indexed A elements are broadcast and processed similarly using {\tt even.indexed\_to\_f32} (lines~15–18), reusing the same accumulator registers—ensuring no spills despite two FMA passes. 
Listing~\ref{code:oddeven} shows the AVX2 packed-operation VNNI nanokernel assembly for the blocking configuration in Figure~\ref{fig:resgister-block}(d).
Lines~1–3 load the packed BF16 odd-indexed elements and convert them to FP32 in $\%ymm12$–$\%ymm14$, while line~4 broadcasts the 0th odd A element into $\%ymm15$.
The \textit{odd} FMAs accumulate results in $\%ymm0$–$\%ymm11$; lines~10 and above repeat the same sequence for the \textit{even} FMAs using the same accumulators.

\begin{lstlisting}[style=customC, basicstyle=\ttfamily\scriptsize, label=code:oddeven, float, caption=BF16 AVX2 VNNI packed  nanokernels., belowskip=-16pt]
vcvtneobf162ps (%rcx),%ymm12//odd load of 0th B chunk
vcvtneobf162ps 32(%rcx),%ymm13//odd load of 1 B chunk
vcvtneobf162ps 64(%rcx),%ymm14//odd load of 2 B chunk
vbcstnebf162ps 12(%rsi),%ymm15//bcst odd 0 ele to F32 
vfmadd231ps   %ymm12, %ymm15, %ymm1 //0th odd fma
vfmadd231ps   %ymm13, %ymm15, %ymm4 //1st odd fma
vfmadd231ps   %ymm15, %ymm14, %ymm8 //2nd odd fma
vbcstnebf162ps -62(%rsi),%ymm15//bcst 1st ele to F32
... // assembly skipped for space
vcvtneebf162ps (%rcx),%ymm12//even load of 0th B chunk
vcvtneebf162ps 32(%rcx),%ymm13//even ld of 1st B chunk
vcvtneebf162ps 64(%rcx),%ymm14//even ld of 2nd B chunk
vbcstnebf162ps 18(%rsi),%ymm15//bcst even 0 ele to F32
vfmadd231ps   %ymm12, %ymm15, %ymm1//0th even fma
vfmadd231ps   %ymm13, %ymm15, %ymm4//1st even fma
vfmadd231ps   %ymm15, %ymm14, %ymm8//2nd even fma
\end{lstlisting}

\paragraph{Flat Layout.}
For flat inputs, VNNI packing is unnecessary since computation is emulated using {\tt vector.fma}.
We exploit AVX2 BF16 packed operations by broadcasting each A$_{i=0..M}$ element (BF16) into {\tt vector<8×f32>} with {\tt bcst\_to\_f32}.
Then, for each B subcolumn chunk, both \textit{even} and \textit{odd} indexed elements are loaded and used in FMA computations (Algorithm~\ref{alg:bf16+splat}, lines~14–19).

\subsubsection{BF16 Fallback}
When the target machine lacks both native BF16 support and AVX2 packed operations, we emulate nanokernel generation similar to the \textit{AVX2 Packed Operations} case.
The A and B input BF16 elements are loaded and converted to FP32 using generic operations such as {\tt bitcast}, {\tt andi}, {\tt select}, and {\tt left\_shift}, followed by fused multiply–add (FMA) computation as shown in Algorithm~\ref{alg:bf16+vnni} (lines~20–31).
As in the Flat AVX2-packed case, the input matrices need not be VNNI-packed; the same sequence of operations is applied to load and convert the A and B BF16 elements to FP32, which are then used directly in the FMA computation (Algorithm~\ref{alg:bf16+splat}, lines~21–27).

% \paragraph{Summarization.}
% Table~\ref{table:mlir-op} summarizes the mapping between MLIR dialect operations and target architectures used in our nanokernel code generation process.
% Each backend employs the most suitable operations for its instruction set: AMX uses tile operations for block loads and fused multiplication; AVX512 leverages vector load/store with the {\tt avx512.dot} intrinsic; AVX2 uses indexed conversion with {\tt vector.fma} for emulation; and generic targets without BF16 support rely on type-conversion primitives combined with {\tt vector.fma}.
% For FP32 workloads, the same vector operations are reused across all ISAs, ensuring portability and consistency in code generation.

For brevity, we present only the assembly of the generated nanokernels here.
The appendix provides detailed compilation steps illustrating how BRGEMM operations are lowered to nanokernels across different target architectures, along with the corresponding MLIR dialect operations used in each case.
Our custom nanokernel generation pass has been integrated into the TPP-MLIR compiler and can be invoked via a dedicated compiler flag, making it readily accessible to all users.
We are currently in the process of upstreaming our code generation and compilation scheme into the LLVM/MLIR compiler, which will enable users to leverage the upstream implementation directly in the future.

% \begin{table}
% \centering
% \begin{tabular}{l |l |l |l} 
%  \hline
%  \textbf{Type} & \textbf{Target} & \textbf{Matrix Data Op} & \textbf{Matmul} \\
%  \hline
%  \multirow{8}{*}{BF16} 
%  & AMX & {\tt amx.tile\_load} & {\tt amx.tile\_mulf} \\
%  & & {\tt amx.tile\_store} &  \\
% \cline{2-4}
%  & AVX512 & {\tt vector.load} & {\tt avx512.dot}{*} \\
%  & \_bf16 & {\tt vector.store} &  \\
%  \cline{2-4}
%    & AVX2\_ & {\tt even.indx\_to\_f32}{*} & {\tt vector.fma} \\
%  & Pk-Op & {\tt odd.indx\_to\_f32}{*} &  \\
%  \cline{2-4}
%   & any ISA  & {\tt vector.load/store} & {\tt vector.fma} \\
%   &  & {\tt bitcast,shli,andi} &  \\
% \hline
% \multirow{2}{*}{FP32}
%  & any ISA & {\tt vector.load} & {\tt vector.fma} \\
%  & & {\tt vector.store} &  \\
%  \hline
% \end{tabular}
% \caption{Mapping of MLIR dialect operations to target machines used in our nanokernel code generation process. {*} - {\tt x86vector} dialect operations.}
% \label{table:mlir-op}
% \vspace{-0.9cm}
% \end{table}

\section{Discussion}
\label{sec:dis}
We now discuss some interesting underlying points about
our proposed compilation scheme.

% \noindent(a) \textbf{Novelty:} The key contribution (novelty) of this work lies in demonstrating that a compiler can automatically generate highly optimized nanokernels without relying on external libraries. To the best of our knowledge, no existing compiler framework has attempted and achieved performance comparable to specialized external libraries through fully compiler-generated nanokernels.

% Our approach provides a clean and self-contained compilation flow that removes dependencies on external libraries and avoids the need for runtime JIT compilation. We integrate the nanokernel generation techniques directly at the IR level within the compiler.

\noindent(a) \textbf{Determining tile sizes.} The main goal of this work is to demonstrate that a compiler can generate optimized kernels comparable to those produced by external libraries. Our nanokernel generation assumes that the input matmul %(only batch or batch-reduce variants) 
is already register-tiled, ensuring that the generated nanokernels are highly optimized. 

We relied on the upstream MLIR pass for register-level tiling by providing the M, and N tile sizes as inputs. Automatic determination of optimal tile sizes by inferring the underlying machine architecture is a separate line of work that is currently in progress.
For the time being, in our experiments, we manually explored different tile sizes for the M and N dimensions and reported the configurations that achieved the best performance.

\noindent(b) \textbf{Portability to other architectures.}
For other architectures (such as {\tt aarch64} or {\tt RISC-V}) or ISAs, our technique naturally extends to those with corresponding MLIR dialects. FP32 nanokernel generation remains unaffected as it relies on generic, architecture-agnostic instructions. For BF16, if a target architecture introduces new ISA support, our custom pass can be extended accordingly—similar to adaptations in external libraries; otherwise, we fall back to generic instructions to ensure correctness.

\noindent(c) \textbf{Support for other packed types.} We can extend or follow  the BF16 nanokernel technique to support other packed types like Int8 and F16. For example, Intel's ArrowLake machine supports  {\tt avx.dot.i8} Int8 operation (like {\tt avx512.dot} BF16 operation) to perform dot product. We could follow the {\tt DotBF16Op} technique to load the A subrow and B subcolumn 
and perform {\tt avx.dot.i8} operation instead of {\tt avx512.dot}. Similarly, with the AVX2 Packed Instruction set we can load or broadcast F16 type elements to F32 and perform FMAs.
Currently, we are generalizing the BF16 nanokernel technique independent of target architecture and planning to upstream both FP32 and BF16 techniques into MLIR/LLVM compiler infrastructure. Afterwards, we plan to extend the nanokernel generation for Int8 type along with quantization support as a future work.

\noindent (d) \textbf{Generality to other matmul operations.}
Throughout this paper, we have used {\tt batch\_reduce.matmul} as the primary example to illustrate nanokernel generation. The same techniques can be readily extended to other {\tt batch.matmul} and {\tt matmul} operations.

\section{Experiment and Evaluation}
\label{sec:exper}

We evaluated the proposed nanokernel generation technique across three recent Intel ISAs. The test platforms are: % as follows:
% \begin{itemize}
% \item[1.] Dual-socket Xeon Platinum 8592+ (Emerald Rapids, EMR) with 2$\times$64 cores, 1024 GB DDR5@5600/DDR4@4400, supporting AMX, AVX-512, and AVX2 instruction sets,
% \item[2.] Dual-socket Xeon 6780E (Sierra Forest, SRF) with 2$\times$144 E-cores, 1024 GB DDR5@6400, supporting AVX2 and AVX2 BF16 packed instruction sets,
% \item[3.] Single-socket Intel Core Ultra 9 285K (Arrow Lake, ARL) with 8 P-cores and 16 E-cores, 96 GB DDR5@7200, supporting AVX2 and AVX2 BF16 packed instruction sets.
% \end{itemize}

%\begin{itemize}
1. Dual-socket Xeon Platinum 8592+ (\textbf{Emerald Rapids, EMR}) with 2$\times$64 cores, 1024 GB DDR5@5600/DDR4@4400, supporting AMX, AVX-512, and AVX2 instruction sets,

2. Dual-socket Xeon 6780E (\textbf{Sierra Forest, SRF}) with 2$\times$144 E-cores, 1024 GB DDR5@6400, supporting AVX2 and AVX2 BF16 packed instruction sets,

3. Single-socket Intel Core Ultra 9 285K (\textbf{Arrow Lake, ARL}) with 8 P-cores and 16 E-cores, 96 GB DDR5@7200, supporting AVX2 and AVX2 BF16 packed instruction sets.
%\end{itemize}

\begin{table}
  \centering
  \begin{tabular}{l|cc|cc|cc}
    \hline
    \textbf{Matrix} &
    \multicolumn{2}{c|}{\textbf{FP32}} &
    \multicolumn{2}{c|}{\textbf{BF16-VNNI}} & 
     \multicolumn{2}{c}{\textbf{BF16-Flat}}\\
    %\cmidrule(lr){2-7} \cmidrule(lr){4-5} \cmidrule(lr){6-7}
    \cline{2-7}
    %\hline
     \textbf{size} & \textbf{EMR} & \textbf{SRF} & \textbf{EMR} & \textbf{SRF} & \textbf{EMR} & \textbf{SRF} \\
    \hline
1024 & 91\% & 93\% & 73\% & 96\% & 91\% & 113\% \\
2048 & 93\% & 91\% & 83\% & 95\% & 95\% & 113\% \\
4096 & 94\% & 93\% & 91\% & 93\% & 93\% & 112\% \\
    \hline
  \end{tabular}

    \caption{\% Geometric mean performance comparison with libxsmm on EMR and SRF across different thread counts and matrix sizes.}
  \label{tab:perf1}
  \vspace{-0.4cm}
\end{table}

\begin{table}
  \centering
  \begin{tabular}{l|cc|cc|cc}
    \hline
    \textbf{Matrix-} &
    \multicolumn{2}{c|}{\textbf{FP32}} &
    \multicolumn{2}{c|}{\textbf{BF16-VNNI}} & 
     \multicolumn{2}{c}{\textbf{BF16-Flat}}\\
    \cline{2-7}
     \textbf{size} & \textbf{P(C)} & \textbf{E(C)} & \textbf{P(C)} & \textbf{E(C)} & \textbf{P(C)} & \textbf{E(C)} \\
    \hline
1024 & 91\% &  93\% &  90\% &  98\% &  132\% &  126\% \\
2048  & 87\% &  87\% &  89\% &  94\% &  132\% &  123\% \\
4096 & 86\% &  85\% &  89\% &  92\% &  131\% &  119\% \\
    \hline
  \end{tabular}
  \caption{\% Geometric mean performance comparison with libxsmm on ARL P and E cores across different thread counts and matrix sizes.}
  \label{tab:perf2}
  \vspace{-0.4cm}
\end{table}

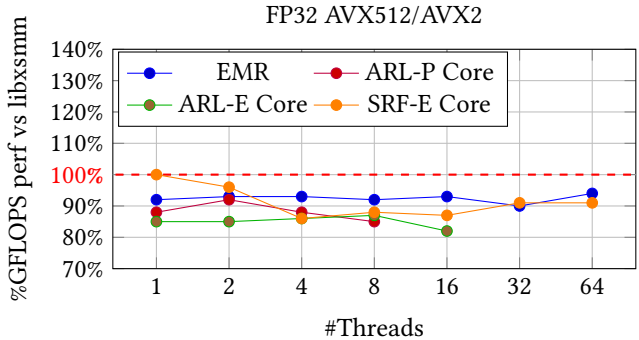
\begin{figure}
\centering
\pgfplotstableread[col sep=comma]{plots/input_gemm_perf2.txt}{\datatable}

\begin{tikzpicture}
\begin{axis}[
    title={FP32 AVX512/AVX2},
    xlabel={},
    ylabel={},
    xticklabels from table={\datatable}{threads},
    xtick=data,
    ymin=70, ymax=140,
    width=.48\textwidth,
    height=4.5cm,
    ytick={70,80,90,110,120,130,140},
    yticklabel=\pgfmathprintnumber{\tick}\%,
    ylabel={\%GFLOPS perf vs libxsmm},
    grid=major,
    xlabel={\#Threads},
    legend style={at={(0.01,0.99)}, anchor=north west,
        legend columns=2,
        %font=\footnotesize, legend cell align=left,
    },
    extra y ticks={100},
    extra y tick style={y tick label style={red, font=\bfseries}},
]

\draw[color=red, thick, dashed] (axis cs:-1,100) -- (axis cs:45,100);

% Example data series (replace with real values)
\addplot+[mark=*,blue] table[x expr=\coordindex, y=emr-f32] {\datatable};
\addlegendentry{EMR}

\addplot+[mark=*,purple] table[x expr=\coordindex, y=arl-p-f32] {\datatable};
\addlegendentry{ARL-P Core}

\addplot+[mark=*,green!70!black] table[x expr=\coordindex, y=arl-e-f32] {\datatable};
\addlegendentry{ARL-E Core}

\addplot+[mark=*,orange] table[x expr=\coordindex, y=srf-f32] {\datatable};
\addlegendentry{SRF-E Core}

\end{axis}
\end{tikzpicture}
\vspace{-0.4cm}
    \caption{\% GFLOPS performance of MLP FP32 AVX nanokernels vs libxsmm on EMR, SRF, and ARL machines.}%  {\tt x-axis} is \#Threads; {\tt y-axis} is GFLOPS.}
  \label{fig:f32}
  \vspace{-0.3cm}
\end{figure}

We implemented our nanokernel lowering techniques within the TPP-MLIR~\cite{tppmlir-git} framework.
All experiments were performed using the latest \texttt{main} branch of the LLVM compiler infrastructure~\cite{llvm-git}, employing both \texttt{clang} and \texttt{gcc}.
We extended the TPP-MLIR pipeline to integrate our nanokernel generation pass, removing the dependency on the libxsmm library and its runtime JIT, thereby enabling a unified ahead-of-time (AOT) compilation path for BRGEMM kernels.

\textbf{Multilayer Perceptron (MLP).}
We evaluated our approach using the MLP kernel (batch-reduce matmul with bias and ReLU) with a batch size of 512 and matrix sizes 1024, 2048, and 4096.
Using TPP-MLIR’s \texttt{mlir-gen} binary, we generated MLP kernels for each configuration.
Tile sizes ($M$, $N$) were selected based on the target instruction set—AMX, AVX512, AVX2, and AVX2 packed.
Performance was measured in GFLOPS and compared against libxsmm, reporting the geometric mean of five runs as a percentage of libxsmm.

\textbf{Output Verification.}
Numerical correctness was verified using the \texttt{tpp-run} binary with a fixed random seed to initialize matrices $A$, $B$, and $C$.
The resulting $C$ matrix was compared against libxsmm outputs, confirming that all nanokernels were both \emph{sound} and \emph{complete}.

Tables~\ref{tab:perf1} and~\ref{tab:perf2} summarize the geometric mean performance, reported as a percentage relative to libxsmm, on EMR, ARL, and SRF systems. Experiments were scaled up to 64 threads for EMR and SRF, up to 8 threads for ARL P-cores, and up to 16 threads for ARL E-cores. Results are presented for matrix sizes 1024, 2048, and 4096, each with batch size 512. For every configuration, we tuned $M$ and $N$ tile sizes to match libxsmm’s performance envelope. The optimal tile sizes were determined to be $M{=}8$, $N{=}32$ for AVX-512; $M{=}2$, $N{=}32$ for AVX2; and $M{=}32$, $N{=}32$ for AMX.

As shown in the tables, our generated nanokernels achieve up to 90\% of libxsmm’s performance across most configurations. For BF16 flat layouts, the nanokernels even outperform libxsmm on both SRF and ARL systems. The following section provides a detailed breakdown of performance trends across varying thread counts for selected matrix sizes. Furthermore, our nanokernels fuse the \textit{bias} and \textit{ReLU} operations before storing the final accumulation into the C matrix, contributing approximately 5\% to the overall performance.

\subsection{Performance of FP32 nanokernels}

Figure~\ref{fig:f32} shows the relative GFLOPS of our FP32 nanokernels for AVX-512 and AVX2 compared to \texttt{libxsmm} across EMR, SRF, and ARL systems for a matrix size of 2048. The 100\% \texttt{libxsmm} baseline is marked by a dashed red line. For ARL, results are reported separately for performance (P) and efficiency (E) cores.
Our FP32 nanokernels closely match \texttt{libxsmm} in both single- and multi-threaded modes, achieving on average (geometric mean) 93\% and 91\% of \texttt{libxsmm} performance on EMR and SRF, respectively, and 88\% on P-cores and 85\% on E-cores of ARL.

\begin{figure}
\centering
\pgfplotstableread[col sep=comma]{plots/input_gemm_perf2.txt}{\datatable}

\begin{tikzpicture}
\begin{axis}[
    title={BF16 AMX},
    xlabel={},
    ylabel={},
    xticklabels from table={\datatable}{threads},
    xtick=data,
    ymin=70, ymax=140,
    width=.48\textwidth,
    height=4.5cm,
    ytick={70,80,90,110,120,130,140},
    yticklabel=\pgfmathprintnumber{\tick}\%,
    ylabel={\%GFLOPS perf vs libxsmm},
    grid=major,
    xlabel={\#Threads},
    legend style={at={(0.01,0.99)}, anchor=north west,
        legend columns=3,
        %font=\footnotesize, legend cell align=left,
    },
    extra y ticks={100},
    extra y tick style={y tick label style={red, font=\bfseries}},
]

\draw[color=red, thick, dashed] (axis cs:-1,100) -- (axis cs:45,100);

% Example data series (replace with real values)
\addplot+[mark=*,blue] table[x expr=\coordindex, y=amx-vnni] {\datatable};
\addlegendentry{VNNI}

\addplot+[mark=*,purple] table[x expr=\coordindex, y=amx-splat] {\datatable};
\addlegendentry{Flat}

% \addplot+[mark=*,green!70!black] table[x expr=\coordindex, y=arl-e-f32] {\datatable};
% \addlegendentry{ARL-E Core}

\end{axis}
\end{tikzpicture}
\vspace{-0.6cm}
    \caption{\% GFLOPS performance of MLP BF16 VNNI/Flat AMX nanokernels vs libxsmm on an EMR machine.}%  {\tt x-axis} is \#Threads; {\tt y-axis} is GFLOPS.}
  \label{fig:amx}
  \vspace{-0.3cm}
\end{figure}
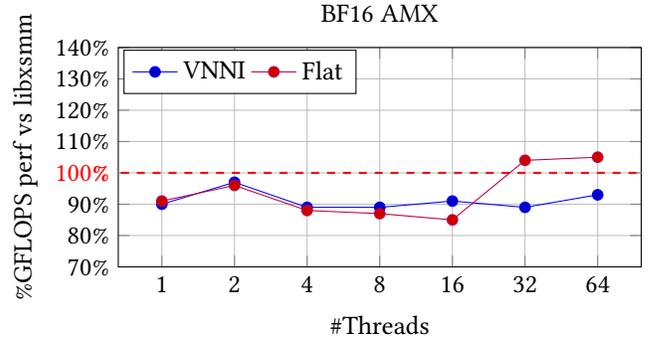

\begin{figure}
\centering
\pgfplotstableread[col sep=comma]{plots/input_gemm_perf2.txt}{\datatable}

\begin{tikzpicture}

\begin{axis}[
    title={BF16 AVX2},
    xlabel={},
    ylabel={},
    xshift=9cm,
    xticklabels from table={\datatable}{threads},
    xtick=data,
    ymin=80, ymax=160,
    width=.48\textwidth,
    height=5.0cm,
    ytick={80,90,110,120,130,140,150,160},
    yticklabel=\pgfmathprintnumber{\tick}\%,
    ylabel={\%GFLOPS perf vs libxsmm},
    grid=major,
    xlabel={\#Threads},
    legend style={at={(0.01,0.99)}, anchor=north west,
        legend columns=3,
        %font=\footnotesize, legend cell align=left,
    },
    extra y ticks={100},
    extra y tick style={y tick label style={red, font=\bfseries}},
]

\draw[color=red, thick, dashed] (axis cs:-1,100) -- (axis cs:45,100);

\addplot+[mark=*,green!70!black] table[x expr=\coordindex, y=srf-bf16-v] {\datatable};
\addlegendentry{V-SRF(E)}

\addplot+[mark=*,purple] table[x expr=\coordindex, y=arl-e-vnni] {\datatable};
\addlegendentry{V-ARL(E)}

\addplot+[mark=*,blue] table[x expr=\coordindex, y=arl-p-vnni] {\datatable};
\addlegendentry{V-ARL(P)}

\addplot+[mark=*,cyan] table[x expr=\coordindex, y=srf-bf16-s] {\datatable};
\addlegendentry{F-SRF(E)}

\addplot+[mark=*,green] table[x expr=\coordindex, y=arl-e-splat] {\datatable};
\addlegendentry{F-ARL(E)}

\addplot+[mark=*,orange] table[x expr=\coordindex, y=arl-p-splat] {\datatable};
\addlegendentry{F-ARL(P)}

\end{axis}

\end{tikzpicture}
\vspace{-0.6cm}
    \caption{\% GFLOPS performance of MLP BF16 VNNI/Flat AVX2 packed operations nanokernels vs libxsmm  on SRF and ARL machines. V-SRF(E): VNNI-SRF (E Cores).}%  {\tt x-axis} is \#Threads; {\tt y-axis} is GFLOPS.}
  \label{fig:avv2-bf16}
  \vspace{-0.3cm}
\end{figure}
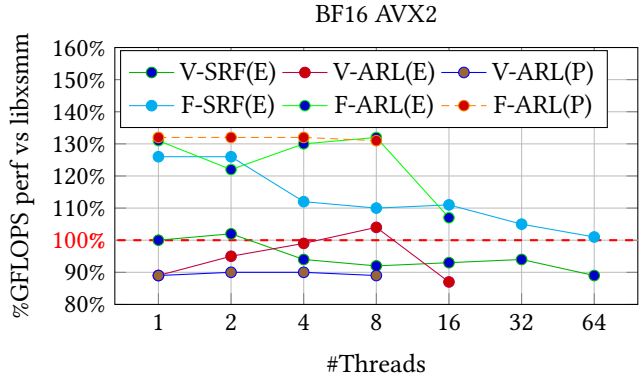

% \subsection{Performance of BF16 AMX nanokernels}

% We capture \% GFLOPS performance for the same BRGEMM kernel, but with layer size increased to 2048 to compensate for the high compute power of AMX instruction set. With blocking size: M:32; N:32; K:32 the \% GFLOPS performance of the AMX nanokernel is shown in figure~\ref{fig:amx}. The AMX Flat nanokernels on {\tt 64} OMP threads out-performance LIBXSMM by +5\%.
% In a geometric mean, the nanokernels reach 90\% and 93\% performance of libxsmm for VNNI and Flat layouts, respectively.
% AMX can perform far more calculations than FP32, which makes memory bandwidth the main bottleneck. If data loads and stores are misaligned, performance can drop to one-fourth of peak, causing around a 30\% slowdown overall.

\subsection{Performance of BF16 AMX nanokernels}
% We report the relative \% GFLOPS performance of our BF16 AMX nanokernels on the same MLP kernel, for matrix size 4096  in Figure~\ref{fig:amx}.
% The results demonstrate that AMX Flat nanokernels executed with 32/64 OpenMP threads outperform libxsmm by approximately +5\%. On average (geometric mean across threads), the nanokernels achieve 91\% and 93\% of libxsmm performance for VNNI and Flat layouts, respectively.
% As expected, AMX can sustain significantly higher compute throughput than FP32; however, this also makes memory bandwidth the primary performance bottleneck. In particular, misaligned data loads and stores can reduce effective throughput to one-fourth of peak, leading to an overall slowdown of around 30\%.
% 
We report the relative BF16 AMX nanokernel performance (in \% GFLOPS) for the 4096-sized MLP kernel in Figure~\ref{fig:amx}. AMX Flat nanokernels executed with 32 or 64 OpenMP threads outperform libxsmm by about 5\%. On average (geometric mean across threads), they reach 91\% and 93\% of libxsmm’s performance for VNNI and Flat layouts, respectively. While AMX delivers much higher compute throughput than FP32, memory bandwidth remains the main bottleneck—misaligned loads and stores can cut effective throughput to one-fourth of peak, causing up to 30\% slowdown.

%\textcolor{red}{TODO: write about the memory effect issue for amx code at llvm backend}

% \begin{figure}
% \centering
% \input{plots/gemm-dp-bf16}
%     \caption{\% GFLOPS performance of BRGEMM bf16 vnni/Flat AVX512\_BF16 nanokernels compared with libxsmm code on EMR and Zen5 machines.}
%   \label{fig:dpbf16}
% \end{figure}

\subsection{Performance of BF16 AVX nanokernels}

% \subsubsection{DotBF16Op nanokernels} We are awaiting for the release of Intel's NovaLake to test our DotBF16Op nanokernels as the NovaLake contains  AMX512\_BF16 vector instruction set that supports BF16 dot-product computation.

% Figure~\ref{fig:dpbf16}, shows the \% GFLOPS performance of BRGEMM bf16 vnni/Flat AVX512\_BF16 nanokernel on an EMR and Zen5 machine for the same BRGEMM kernel with layers 1024 and blocking size: M:8; N:32; K:2. The plot clearly shows that our nanokernel could match the performance of libxsmm for the vnni layout and performs better than libxsmm for Flat layout. In a geometric mean, the nanokernels reach 100\% and 115\% performance of libxsmm for the vnni and Flat layouts, respectively.
% We investigated the reason for better performance than libxsmm in Flat layouts and found that libxsmm uses {\tt vpermt2pd} instruction to shuffle the B subcolumn, while the nanokernel uses {\tt vpunpcklwd} and {\tt vpunpckhwd}. The latency of {\tt vpunpcklwd/hwd} instruction is low compared to {\tt vpermt2pd} because they operate at 16 bit words which is best suited for BF16 type.

\subsubsection{AVX2 Packed Instructions}
% We evaluated BF16 AVX2 packed instruction nanokernel lowering on the Intel SRF and ARL machines. Figure~\ref{fig:avv2-bf16} reports the percentage GFLOPS performance of our generated nanokernels compared to libxsmm for matrix size 2048, with results shown separately for P-cores and E-cores.
% On the SRF machine, the VNNI nanokernel achieves 95\% of libxsmm performance. On the ARL machine, the VNNI nanokernel reaches 89\% on P-cores and 94\% on E-cores. In contrast, the Flat nanokernel consistently outperforms libxsmm, achieving 113\% on SRF, 132\% on ARL P-cores, and 122\% on ARL E-cores.
% This performance gap arises because libxsmm internally converts the $B$ subcolumns from Flat to VNNI-packed format, whereas our nanokernels directly exploit AVX2 packed operations. These results highlight the advantage of compiler-integrated nanokernel generation, which enables systematic selection of operations that best match the target architecture and workload.
%We evaluated BF16 AVX2-packed nanokernel lowering on Intel SRF and ARL systems. 
Figure~\ref{fig:avv2-bf16} shows relative GFLOPS for matrix size 2048, reported separately for P-cores and E-cores. On SRF, the VNNI nanokernel attains 95\% of libxsmm’s performance; on ARL, it reaches 89\% on P-cores and 94\% on E-cores. In contrast, the Flat nanokernel outperforms libxsmm, achieving 113\% on SRF, 132\% on ARL P-cores, and 122\% on ARL E-cores. The performance gap stems from libxsmm’s internal conversion of $B$ subcolumns from Flat to VNNI format, whereas our approach directly leverages AVX2 packed operations—demonstrating the advantage of compiler-integrated nanokernel generation in aligning operations with target architectures and workloads. 
Both the VNNI-packed and Flat layouts exhibit a performance dip at 16 threads on ARL E-cores. Initial analysis indicates that this issue is related to the choice of {\tt --parallel-task-grid}. We are currently further investigating with PARLOOPER~\cite{parlooper} to address and resolve this behavior.

\subsubsection{DotBF16Op Nanokernels}
% We are awaiting the release of Intel’s Nova Lake processors to evaluate our \texttt{DotBF16Op} nanokernels. Nova Lake introduces the AVX512\_BF16 vector instruction set, which provides native support for BF16 dot-product computation. Based on the trends observed in our FP32 and BF16 AMX evaluations, we expect these nanokernels to achieve competitive performance relative to libxsmm, while further benefiting from reduced memory bandwidth pressure inherent to BF16. A detailed evaluation of this class of nanokernels is left as part of our future work.
We plan to evaluate our \texttt{DotBF16Op} nanokernels on Intel’s upcoming Nova Lake processors, which add AVX512\_BF16 instructions for native BF16 dot products. Based on FP32 and BF16 AMX/AVX2 trends, we expect competitive performance aganist libxsmm.

\textbf{Summary.} The compiler-generated nanokernels match or surpass libxsmm across FP32, BF16 AMX, and BF16 AVX2 benchmarks. FP32 kernels sustain 88–97\% of libxsmm on EMR, SRF, and ARL. BF16 AMX averages 90–93\%, with Flat layouts up to +5\% higher. BF16 AVX2 Flat variants outperform by 18–33\%. Overall, compiler-integrated nanokernels rival hand-tuned libraries while efficiently exploiting ISA features for unified BRGEMM generation.

\section{Related Work}
\label{sec:related}

Bondhugula ~\cite{hpc} demonstrated that MLIR, with its affine dialect and polyhedral utilities, can generate high-performance GEMM code rivaling hand-tuned libraries through modular optimization passes like tiling, packing, and vectorization. Another recent work \cite{exo} outlines a DSL-based microkernel generation approach, but it requires manual setup and lacks production-level portability. Braedy et al. \cite{intrinsic} achieved near-peak performance using a compiler-only LLVM approach, but its tight coupling to LLVM and focus on POWER CPUs limits extensibility and portability, unlike MLIR’s vector dialect which targets diverse hardware. Our work addresses these gaps by leveraging MLIR’s modular design and target abstraction (e.g., {\tt x86Vector} dialect) to support Intel AVX2, AVX512, and AMX, enabling scalable intrinsic lowering and ML framework integration. Gareev et al. \cite{tc} used Polly and affine transformations to optimize tensor contractions, achieving strong performance but requiring manual tuning and lacking ML framework integration. Shah et al. \cite{mirokernels} proposed macro-based vector intrinsic abstractions for portable microkernels, but their approach demands architectural expertise and manual tuning. In contrast, our system automates nanokernel generation using MLIR’s vector dialects and custom lowering, abstracting hardware details while enabling fine-grained control. We support compiler-driven composition of nanokernels into microkernels, facilitating rapid adaptation to new instruction sets like AMX and VNNI. Building on foundational work by Vasilache et al. \cite{composable}, we specialize MLIR pipelines for GEMM-based microkernels, applying register-aware lowering strategies. Overall, our approach extends MLIR’s modular philosophy to deliver scalable, portable, and production-grade kernel generation tailored for modern AI workloads.

% \section{Future Work}
% \label{sec:fw}
% \input{08_future.tex}

\section{Conclusion}
\label{sec:con}
% We presented a technique to automatically generate target-specific nanokernels from contraction operations within the MLIR/LLVM compiler infrastructure. Our approach removes the reliance on external microkernel libraries and runtime JIT compilation, while still delivering competitive performance against state-of-the-art libraries such as libxsmm. In fact, our evaluation shows that compiler-generated nanokernels can consistently achieve 88–133\% of libxsmm performance across recent Intel architectures, underscoring both their competitiveness and their potential to surpass hand-tuned microkernels.
% Although our current implementation targets BRGEMM, the methodology naturally extends to matmul and batch-matmul operations. With emerging processors offering native support for increasingly diverse packed data types, we argue that compiler-integrated nanokernel generation is not only viable but necessary to fully exploit the computational capabilities of modern hardware.

% As future work, we plan to generalize and upstream our technique into the MLIR/LLVM compiler, thereby enabling the broader community to benefit from a unified path for nanokernel generation. This will open the door to further innovations in data layout, tiling, and ISA-specific optimizations, transforming nanokernel generation into a community-driven, extensible, and future-proof component of the compiler stack.

We introduced a compiler-based approach for automatically generating target-specific nanokernels, eliminating the need for external libraries or runtime JIT compilation. Our techniques, instantiated using the MLIR/LLVM infrastructure, demonstrate both competitive performance and the potential to surpass hand-tuned kernels. As hardware continues to evolve with richer data types, compiler-integrated nanokernel generation will be essential to fully exploit modern architectures. Future work will focus on upstreaming this capability into MLIR/LLVM, enabling community-driven extensions in data layout, tiling, and ISA-specific optimizations.


% \begin{acks}
% \end{acks}

\bibliographystyle{ACM-Reference-Format}
\bibliography{sample-base}
\vspace{0.5cm}
\noindent \textbf{Optimization Notice:} Software and workloads used in
performance tests may have been optimized for performance only on
Intel microprocessors.  Performance tests, such as SYSmark and
MobileMark, are measured using specific computer systems,
components, software, operations and functions.  Any change to any
of those factors may cause the results to vary.  You should
consult other information and performance tests to assist you in
fully evaluating your contemplated purchases, including the
performance of that product when combined with other products.
For more information go to \url{http://www.intel.com/performance}.

\noindent Intel, Xeon, and Intel Xeon Phi are trademarks of Intel Corporation in the U.S. and/or other

\newpage
\appendix

\onecolumn

\section{Appendix}

\label{sec:appendix}
Here, we present the generated nanokernels produced by our compilation scheme, 
targeting different machine architectures, data types, and data layouts.

\subsection{FP32 Nanokernel Generation (Any ISA)}

\begin{lstlisting}[style=customC, basicstyle=\ttfamily\scriptsize, label=code:input-fp32, caption=Input FP32 batch-reduce matmul kernel.]
func.func fp32(%arg0: memref<2x4x32xf32>, %arg1: memref<2x32x96xf32>, %arg2: memref<4x96xf32>) -> memref<4x96xf32> {
 linalg.batch_reduce_matmul ins(%arg0,%arg1 : memref<2x4x32xf32>,memref<2x32x96xf32>) outs(%arg2 : memref<4x96xf32>)
 return %arg2 : memref<4x96xf32>
}
\end{lstlisting}

\begin{lstlisting}[style=customC, basicstyle=\ttfamily\scriptsize, label=code:fp32-tiled, caption= The input kernel is register tiled with M:2 N:32 batch:1 and K:1 using upstream MLIR transformations.]
func.func fp32(%arg0: memref<2x4x32xf32>, %arg1: memref<2x32x96xf32>, %arg2: memref<4x96xf32>) -> memref<4x96xf32> {
 %c0 = arith.constant 0 : index
 %c4 = arith.constant 4 : index
 %c2 = arith.constant 2 : index
 %c96 = arith.constant 96 : index
 %c32 = arith.constant 32 : index
 %c1 = arith.constant 1 : index
 scf.for %arg3 = %c0 to %c4 step %c2 // M Loop
  scf.for %arg4 = %c0 to %c96 step %c32 // N Loop
   scf.for %arg5 = %c0 to %c2 step %c1 // batch Loop
    scf.for %arg6 = %c0 to %c32 step %c1 // K Loop
     %subview = memref.subview %arg0[%arg5, %arg3, %arg6] [1, 2, 1] [1, 1, 1] : memref<2x4x32xf32> to memref<1x2x1xf32, strided<[128, 32, 1], offset: ?>>
     %subview_0 = memref.subview %arg1[%arg5, %arg6, %arg4] [1, 1, 32] [1, 1, 1] : memref<2x32x96xf32> to memref<1x1x32xf32, strided<[3072, 96, 1], offset: ?>>
     %subview_1 = memref.subview %arg2[%arg3, %arg4] [2, 32] [1, 1] : memref<4x96xf32> to memref<2x32xf32, strided<[96, 1], offset: ?>>
     linalg.batch_reduce_matmul ins(%subview, %subview_0 : memref<1x2x1xf32, strided<[128, 32, 1], offset: ?>>, memref<1x1x32xf32, strided<[3072, 96, 1], offset: ?>>) outs(%subview_1 : memref<2x32xf32, strided<[96, 1], offset: ?>>)
 return %arg2 : memref<4x96xf32>
}
\end{lstlisting}

\begin{lstlisting}[style=customC, basicstyle=\ttfamily\scriptsize, label=code:fp32-contraction, caption=The register tiled IR then lowered to vector contraction using upstream vectorization pass.]
func.func fp32(%arg0: memref<2x4x32xf32>, %arg1: memref<2x32x96xf32>, %arg2: memref<4x96xf32>) -> memref<4x96xf32> {
 %0 = ub.poison : f32
 %c0 = arith.constant 0 : index
 %c4 = arith.constant 4 : index
 %c2 = arith.constant 2 : index
 %c96 = arith.constant 96 : index
 %c32 = arith.constant 32 : index
 %c1 = arith.constant 1 : index
 scf.for %arg3 = %c0 to %c4 step %c2 {
  scf.for %arg4 = %c0 to %c96 step %c32 {
   %subview = memref.subview %arg2[%arg3, %arg4] [2, 32] [1, 1] : memref<4x96xf32> to memref<2x32xf32, {{.*}}>
   %1 = vector.transfer_read %subview[%c0,%c0], %0{in_bounds=[true,true]} : memref<2x32xf32,{{.*}}>, vector<2x32xf32>
   %2 = scf.for %arg5 = %c0 to %c2 step %c1 iter_args(%arg6 = %1) -> (vector<2x32xf32>) {
    %3 = scf.for %arg7 = %c0 to %c32 step %c1 iter_args(%arg8 = %arg6) -> (vector<2x32xf32>) {
     %sub_0 = memref.subview %arg0[%arg5,%arg3,%arg7][1,2,1][1,1,1]:memref<2x4x32xf32> to memref<1x2x1xf32,{{.*}}>
     %sub_1 = memref.subview %arg1[%arg5,%arg7,%arg4][1,1,32][1,1,1]:memref<2x32x96xf32> to memref<1x1x32xf32,{{.*}}>
     %4 = vector.transfer_read %sub_0[%c0,%c0,%c0], %0 {in_bounds = [true,true,true]} : memref<1x2x1xf32, {{.*}}>, vector<1x2x1xf32>
     %5 = vector.transfer_read %sub_1[%c0,%c0,%c0], %0 {in_bounds = [true,true,true]} : memref<1x1x32xf32, {{.*}}>, vector<1x1x32xf32>
     %6 = vector.contract {indexing_maps=[#map,#map1,#map2], iterator_types = ["reduction","parallel","parallel","reduction"], kind=#vector.kind<add>} %4, %5, %arg8 : vector<1x2x1xf32>, vector<1x1x32xf32> into vector<2x32xf32>
     scf.yield %6 : vector<2x32xf32>
    }
    scf.yield %3 : vector<2x32xf32>
   }
   vector.transfer_write %2, %subview[%c0,%c0] {in_bounds = [true,true]} : vector<2x32xf32>, memref<2x32xf32, {{.*}}>
  }
 }
 return %arg2 : memref<4x96xf32>
}
\end{lstlisting}

\begin{lstlisting}[style=customC, basicstyle=\ttfamily\scriptsize, label=code:fp32-nano, caption=Our nanokernel generation technique accepts tiled {\tt vector.contract} as input and emits  AVX-512 target-specific FP32 nanokernel using a sequence of vector FMAs. For AVX2 targets - the same code generation 
pattern is applied with a reduced vector length of 8. Apart from the vector width - 
the FP32 code generation process remains identical across different ISAs ensuring 
a consistent and portable lowering pipeline.]
func.func fp32(%arg0: memref<2x4x32xf32>, %arg1: memref<2x32x96xf32>, %arg2: memref<4x96xf32>) -> memref<4x96xf32> {
 %c16 = arith.constant 16 : index
 %c0 = arith.constant 0 : index
 %c4 = arith.constant 4 : index
 %c2 = arith.constant 2 : index
 %c96 = arith.constant 96 : index
 %c32 = arith.constant 32 : index
 %c1 = arith.constant 1 : index
 
 scf.for %arg3 = %c0 to %c4 step %c2 { // M Loop
  scf.for %arg4 = %c0 to %c96 step %c32 { // N Loop
   %subview = memref.subview %arg2[%arg3, %arg4] [2, 32] [1, 1] : memref<4x96xf32> to memref<2x32xf32, {{.*}}>

   // Load C matrix into small vectors (accumulators)
   %1 = vector.load %subview[%c0, %c0] : memref<2x32xf32, strided<[96, 1], offset: ?>>, vector<16xf32>
   %2 = vector.load %subview[%c1, %c0] : memref<2x32xf32, strided<[96, 1], offset: ?>>, vector<16xf32>
   %3 = vector.load %subview[%c0, %c16] : memref<2x32xf32, strided<[96, 1], offset: ?>>, vector<16xf32>
   %4 = vector.load %subview[%c1, %c16] : memref<2x32xf32, strided<[96, 1], offset: ?>>, vector<16xf32>

   // Passing the accumulator as iter args ensure it remains in vector register across the partial GEMMs
   %5:4 = scf.for %arg5 = %c0 to %c2 step %c1 iter_args(%arg6 = %1, %arg7 = %2, %arg8 = %3, %arg9 = %4) -> (vector<16xf32>, vector<16xf32>, vector<16xf32>, vector<16xf32>) { // batch-reduce Loop
    %6:4 = scf.for %arg10 = %c0 to %c32 step %c1 iter_args(%arg11 = %arg6, %arg12 = %arg7, %arg13 = %arg8, %arg14 = %arg9) -> (vector<16xf32>, vector<16xf32>, vector<16xf32>, vector<16xf32>) { // K Loop
    
     %sub_0 = memref.subview %arg0[%arg5,%arg3,%arg10][1,2,1][1,1,1]:memref<2x4x32xf32> to memref<1x2x1xf32,{{.*}}>
     %sub_1=memref.subview %arg1[%arg5,%arg10,%arg4][1,1,32][1,1,1]:memref<2x32x96xf32> to memref<1x1x32xf32,{{.*}}>

     // Load B subchunks
     %7 = vector.load %sub_1[%c0, %c0, %c0] : memref<1x1x32xf32, {{.*}}>, vector<16xf32>
     %8 = vector.load %sub_1[%c0, %c0, %c16] : memref<1x1x32xf32, {{.*}}>, vector<16xf32>

     // Load 0th element of A subrow, broadcast and perform FMA
     %9 = vector.load %sub_0[%c0, %c0, %c0] : memref<1x2x1xf32, {{.*}}>, vector<1xf32>
     %10 = vector.broadcast %9 : vector<1xf32> to vector<16xf32>
     %11 = vector.fma %10, %7, %arg11 : vector<16xf32>
     %12 = vector.fma %10, %8, %arg13 : vector<16xf32>

     // Load 1st element of A subrow, broadcast and perform FMA
     %13 = vector.load %sub_0[%c0, %c1, %c0] : memref<1x2x1xf32, {{.*}}>, vector<1xf32>
     %14 = vector.broadcast %13 : vector<1xf32> to vector<16xf32>
     %15 = vector.fma %14, %7, %arg12 : vector<16xf32>
     %16 = vector.fma %14, %8, %arg14 : vector<16xf32>
     
     scf.yield %11, %15, %12, %16 : vector<16xf32>, vector<16xf32>, vector<16xf32>, vector<16xf32>
    }
    scf.yield %6#0, %6#1, %6#2, %6#3 : vector<16xf32>, vector<16xf32>, vector<16xf32>, vector<16xf32>
   }

   // Store the final accumulator into C matrix
   vector.store %5#0, %subview[%c0, %c0] : memref<2x32xf32, strided<[96, 1], offset: ?>>, vector<16xf32>
   vector.store %5#1, %subview[%c1, %c0] : memref<2x32xf32, strided<[96, 1], offset: ?>>, vector<16xf32>
   vector.store %5#2, %subview[%c0, %c16] : memref<2x32xf32, strided<[96, 1], offset: ?>>, vector<16xf32>
   vector.store %5#3, %subview[%c1, %c16] : memref<2x32xf32, strided<[96, 1], offset: ?>>, vector<16xf32>
  }
 }
 return %arg2 : memref<4x96xf32>
}
\end{lstlisting}

\newpage
\subsection{BF16 VNNI packed AMX target}
\begin{lstlisting}[style=customC, basicstyle=\ttfamily\scriptsize, label=code:input-bf16-vnni, caption=Input BF16 batch-reduce matmul kernel with VNNI:2 packing.]
func.func bf16(%arg0:memref<2x64x32x2xbf16>,%arg1:memref<2x32x64x2xbf16>,%arg2:memref<64x64xf32>)->memref<64x64xf32>{
 linalg.generic {indexing_maps = [affine_map<(d0, d1, d2, d3, d4) -> (d0, d2, d4, d1)>, affine_map<(d0, d1, d2, d3, d4) -> (d0, d4, d3, d1)>, affine_map<(d0, d1, d2, d3, d4) -> (d2, d3)>], iterator_types = ["reduction", "reduction", "parallel", "parallel", "reduction"]} ins(%arg0, %arg1 : memref<2x64x32x2xbf16>, memref<2x32x64x2xbf16>) outs(%arg2 : memref<64x64xf32>) {
 ^bb0(%in: bf16, %in_1: bf16, %out: f32):
   %a = arith.extf %in : bf16 to f32
   %b = arith.extf %in_1 : bf16 to f32
   %1 = arith.mulf %a, %b : f32
   %2 = arith.addf %out, %1 : f32
   linalg.yield %2 : f32
 }
 return %arg2 : memref<64x64xf32>
}
\end{lstlisting}

%\lstinputlisting[style=customC, basicstyle=\ttfamily\scriptsize, label=code:input, caption=Input FP32 batch-reduce matmul kernel.]{mlir/appendix/bf16-tiled.mlir}

\begin{lstlisting}[style=customC, basicstyle=\ttfamily\scriptsize, label=code:bf16-til-con, caption=The input BF16 kernel is register tiled with M:32 N:32 K:32 and further lowered to vector contraction using upstream MLIR transformations.]
#map = affine_map<(d0, d1, d2, d3, d4) -> (d0, d2, d4, d1)>
#map1 = affine_map<(d0, d1, d2, d3, d4) -> (d0, d4, d3, d1)>
#map2 = affine_map<(d0, d1, d2, d3, d4) -> (d2, d3)>

func.func bf16(%arg0:memref<2x64x32x2xbf16>,%arg1:memref<2x32x64x2xbf16>,%arg2:memref<64x64xf32>)->memref<64x64xf32>{
 %0 = ub.poison : f32
 %1 = ub.poison : bf16
 %c0 = arith.constant 0 : index
 %c64 = arith.constant 64 : index
 %c32 = arith.constant 32 : index
 %c2 = arith.constant 2 : index
 %c1 = arith.constant 1 : index
 %c16 = arith.constant 16 : index
 
 scf.for %arg3 = %c0 to %c64 step %c32 { // M Loop
  scf.for %arg4 = %c0 to %c64 step %c32 { // N Loop
   %subview = memref.subview %arg2[%arg3, %arg4] [32, 32] [1, 1] : memref<64x64xf32> to memref<32x32xf32, {{.*}}>
   %2 = vector.transfer_read %subview[%c0,%c0], %0{in_bounds=[true,true]}:memref<32x32xf32,{{.*}}>, vector<32x32xf32>
   
   %3 = scf.for %arg5 = %c0 to %c2 step %c1 iter_args(%arg6 = %2) -> (vector<32x32xf32>) { // batch Loop
    %4 = scf.for %arg7 = %c0 to %c32 step %c16 iter_args(%arg8 = %arg6) -> (vector<32x32xf32>) { // K Loop
     %subview_0 = memref.subview %arg0[%arg5, %arg3, %arg7, 0] [1, 32, 16, 2] [1, 1, 1, 1] : memref<2x64x32x2xbf16> to memref<1x32x16x2xbf16, {{.*}}>
     %subview_1 = memref.subview %arg1[%arg5, %arg7, %arg4, 0] [1, 16, 32, 2] [1, 1, 1, 1] : memref<2x32x64x2xbf16> to memref<1x16x32x2xbf16, {{.*}}>
     %5 = vector.transfer_read %subview_0[%c0, %c0, %c0, %c0], %1 {in_bounds = [true, true, true, true]} : memref<1x32x16x2xbf16, {{.*}}>, vector<1x32x16x2xbf16>
     %6 = vector.transfer_read %subview_1[%c0, %c0, %c0, %c0], %1 {in_bounds = [true, true, true, true]} : memref<1x16x32x2xbf16, {{.*}}>, vector<1x16x32x2xbf16>
     %7 = vector.contract {indexing_maps = [#map, #map1, #map2], iterator_types = ["reduction", "reduction", "parallel", "parallel", "reduction"], kind = #vector.kind<add>} %5, %6, %arg8 : vector<1x32x16x2xbf16>, vector<1x16x32x2xbf16> into vector<32x32xf32>
     
     scf.yield %7 : vector<32x32xf32>
    }
    scf.yield %4 : vector<32x32xf32>
   }
   vector.transfer_write %3, %subview[%c0,%c0] {in_bounds=[true,true]} : vector<32x32xf32>, memref<32x32xf32, {{.*}}>
  }
 }
 return %arg2 : memref<64x64xf32>
}
\end{lstlisting}

\newpage

\begin{lstlisting}[style=customC, basicstyle=\ttfamily\scriptsize, label=code:bf16-amx, caption=The generated AMX target-specific BF16 VNNI-packed nanokernel IR is produced by our 
proposed compilation technique using appropriately selected operations from MLIR’s 
\texttt{AMX} dialect. This intermediate representation serves as the final lowered 
form before target-specific code emission. The assembly code presented in Section~4 
of the main manuscript is directly generated from this IR. ]
func.func bf16(%arg0:memref<2x64x32x2xbf16>,%arg1:memref<2x32x64x2xbf16>,%arg2:memref<64x64xf32>)->memref<64x64xf32>{
 %c0 = arith.constant 0 : index
 %c64 = arith.constant 64 : index
 %c32 = arith.constant 32 : index
 %c2 = arith.constant 2 : index
 %c1 = arith.constant 1 : index
 %c16 = arith.constant 16 : index
 
 scf.for %arg3 = %c0 to %c64 step %c32 { // M Loop
  scf.for %arg4 = %c0 to %c64 step %c32 { // N Loop
  
   // Load C subpanel as accumulators
   %subview = memref.subview %arg2[%arg3, %arg4] [32, 32] [1, 1] : memref<64x64xf32> to memref<32x32xf32, {{.*}}>
   %0 = amx.tile_load %subview[%c0, %c0] : memref<32x32xf32, {{.*}}> into !amx.tile<16x16xf32>
   %1 = amx.tile_load %subview[%c0, %c16] : memref<32x32xf32, {{.*}}> into !amx.tile<16x16xf32>
   %2 = amx.tile_load %subview[%c16, %c0] : memref<32x32xf32, {{.*}}> into !amx.tile<16x16xf32>
   %3 = amx.tile_load %subview[%c16, %c16] : memref<32x32xf32, {{.*}}> into !amx.tile<16x16xf32>
   
   %4:4 = scf.for %arg5 = %c0 to %c2 step %c1 iter_args(%arg6 = %0, %arg7 = %1, %arg8 = %2, %arg9 = %3) -> (!amx.tile<16x16xf32>, !amx.tile<16x16xf32>, !amx.tile<16x16xf32>, !amx.tile<16x16xf32>) { // batch-reduce Loop
    %5:4 = scf.for %arg10 = %c0 to %c32 step %c16 iter_args(%arg11 = %arg6, %arg12 = %arg7, %arg13 = %arg8, %arg14 = %arg9) -> (!amx.tile<16x16xf32>, !amx.tile<16x16xf32>, !amx.tile<16x16xf32>, !amx.tile<16x16xf32>) { // K loop

     // Load A subpanel
     %sub_0 = memref.subview %arg0[%arg5,%arg3,%arg10,0] [1,32,16,2] [1,1,1,1] : memref<2x64x32x2xbf16> to memref<1x32x16x2xbf16,{{.*}}>
     %cs=memref.collapse_shape %sub_0 [[0],[1],[2,3]]:memref<1x32x16x2xbf16,{{.*}}> into memref<1x32x32xbf16,{{.*}}>
     %6 = amx.tile_load %cs[%c0,%c0,%c0] : memref<1x32x32xbf16,{{.*}}> into !amx.tile<16x32xbf16>   
     %cs_1=memref.collapse_shape %sub_0[[0],[1],[2,3]]:memref<1x32x16x2xbf16,{{.*}}> into memref<1x32x32xbf16,{{.*}}>
     %7 = amx.tile_load %cs_1[%c0,%c16,%c0]:memref<1x32x32xbf16,{{.*}}> into !amx.tile<16x32xbf16>

     // Load B subpanel
     %sub_2 = memref.subview %arg1[%arg5,%arg10,%arg4,0] [1,16,32,2] [1,1,1,1] : memref<2x32x64x2xbf16> to memref<1x16x32x2xbf16,{{.*}}>
     %cs_3=memref.collapse_shape %sub_2[[0],[1],[2,3]]:memref<1x16x32x2xbf16,{{.*}}> into memref<1x16x64xbf16,{{.*}}>
     %8 = amx.tile_load %cs_3[%c0,%c0,%c0] : memref<1x16x64xbf16,{{.*}}> into !amx.tile<16x32xbf16>    
     %cs_4=memref.collapse_shape %sub_2[[0],[1],[2,3]]:memref<1x16x32x2xbf16,{{.*}}> into memref<1x16x64xbf16,{{.*}}>
     %9 = amx.tile_load %cs_4[%c0,%c0,%c32] : memref<1x16x64xbf16,{{.*}}> into !amx.tile<16x32xbf16>

     // AMX tiled dotproduct operations
     %10 = amx.tile_mulf %6, %8, %arg11 : !amx.tile<16x32xbf16>, !amx.tile<16x32xbf16>, !amx.tile<16x16xf32>
     %11 = amx.tile_mulf %6, %9, %arg12 : !amx.tile<16x32xbf16>, !amx.tile<16x32xbf16>, !amx.tile<16x16xf32>
     %12 = amx.tile_mulf %7, %8, %arg13 : !amx.tile<16x32xbf16>, !amx.tile<16x32xbf16>, !amx.tile<16x16xf32>
     %13 = amx.tile_mulf %7, %9, %arg14 : !amx.tile<16x32xbf16>, !amx.tile<16x32xbf16>, !amx.tile<16x16xf32>
     
     scf.yield %10, %11, %12, %13 : !amx.tile<16x16xf32>, !amx.tile<16x16xf32>, !amx.tile<16x16xf32>, !amx.tile<16x16xf32>
    }
    scf.yield %5#0, %5#1, %5#2, %5#3 : !amx.tile<16x16xf32>, !amx.tile<16x16xf32>, !amx.tile<16x16xf32>, !amx.tile<16x16xf32>
   }

   // Store accumulator back into C subpanel
   amx.tile_store %subview[%c0, %c0], %4#0 : memref<32x32xf32, {{.*}}>, !amx.tile<16x16xf32>
   amx.tile_store %subview[%c0, %c16], %4#1 : memref<32x32xf32, {{.*}}>, !amx.tile<16x16xf32>
   amx.tile_store %subview[%c16, %c0], %4#2 : memref<32x32xf32, {{.*}}>, !amx.tile<16x16xf32>
   amx.tile_store %subview[%c16, %c16], %4#3 : memref<32x32xf32, {{.*}}>, !amx.tile<16x16xf32>
  }
 }
 return %arg2 : memref<64x64xf32>
}
\end{lstlisting}

\newpage
\subsection{BF16 VNNI packed AVX-512 DotOp target}

\begin{lstlisting}[style=customC, basicstyle=\ttfamily\scriptsize, label=code:bf16-dp, caption=The AVX-512 BF16 target-specific DotOp-based VNNI-packed nanokernel IR is generated 
using our proposed technique for the BF16 VNNI-packed input shown in 
Listing~\ref{code:input-bf16-vnni}. In the first stage - the input kernel is tiled 
with parameters M:2 N:32 K:2 (VNNI) - after which it is further lowered into a 
\texttt{vector.contract} operation. This transformation enables efficient utilization 
of AVX-512 vector units and leverages the use of {\tt x86vector} dialect's VNNI operation ({\tt avx512.dot}) for high-throughput 
BF16 matrix computations.
]
func.func bf16(%arg0:memref<2x64x32x2xbf16>,%arg1:memref<2x32x64x2xbf16>,%arg2:memref<64x64xf32>)->memref<64x64xf32>{
 %c16 = arith.constant 16 : index
 %c0 = arith.constant 0 : index
 %c64 = arith.constant 64 : index
 %c2 = arith.constant 2 : index
 %c32 = arith.constant 32 : index
 %c1 = arith.constant 1 : index

 scf.for %arg3 = %c0 to %c64 step %c2 { // M Loop
  scf.for %arg4 = %c0 to %c64 step %c32 { // N Loop
   %subview = memref.subview %arg2[%arg3, %arg4] [2, 32] [1, 1] : memref<64x64xf32> to memref<2x32xf32, {{.*}}>

   // Load C matrix into small vectors (accumulators)
   %1 = vector.load %subview[%c0, %c0] : memref<2x32xf32, {{.*}}>, vector<16xf32>
   %2 = vector.load %subview[%c1, %c0] : memref<2x32xf32, {{.*}}>, vector<16xf32>
   %3 = vector.load %subview[%c0, %c16] : memref<2x32xf32, {{.*}}>, vector<16xf32>
   %4 = vector.load %subview[%c1, %c16] : memref<2x32xf32, {{.*}}>, vector<16xf32>

   // Passing the accumulator as iter args ensure it remains in vector register across the partial GEMMs
   %5:4 = scf.for %arg5 = %c0 to %c2 step %c1 iter_args(%arg6 = %1, %arg7 = %2, %arg8 = %3, %arg9 = %4) -> (vector<16xf32>, vector<16xf32>, vector<16xf32>, vector<16xf32>) { // batch-reduce Loop
    %6:4 = scf.for %arg10 = %c0 to %c32 step %c1 iter_args(%arg11 = %arg6, %arg12 = %arg7, %arg13 = %arg8, %arg14 = %arg9) -> (vector<16xf32>, vector<16xf32>, vector<16xf32>, vector<16xf32>) { // K Loop
    
     %sub_0 = memref.subview %arg0[%arg5, %arg3, %arg10, 0] [1, 2, 1, 2] [1, 1, 1, 1] : memref<2x64x32x2xbf16> to memref<1x2x1x2xbf16, {{.*}}>
     %sub_1 = memref.subview %arg1[%arg5, %arg10, %arg4, 0] [1, 1, 32, 2] [1, 1, 1, 1] : memref<2x32x64x2xbf16> to memref<1x1x32x2xbf16, {{.*}}>

     // Load B subchunks
     %7 = vector.load %sub_1[%c0, %c0, %c0, %c0] : memref<1x1x32x2xbf16, {{.*}}>, vector<32xbf16>
     %8 = vector.load %sub_1[%c0, %c0, %c16, %c0] : memref<1x1x32x2xbf16, {{.*}}>, vector<32xbf16>

     // Load the first pair of BF16 elements of A subrow, broadcast and perform BF16 dotproduct
     %9 = vector.load %sub_0[%c0, %c0, %c0, %c0] : memref<1x2x1x2xbf16, {{.*}}>, vector<2xbf16>
     %10 = vector.bitcast %9 : vector<2xbf16> to vector<1xi32>
     %11 = vector.broadcast %10 : vector<1xi32> to vector<16xi32>
     %12 = vector.bitcast %11 : vector<16xi32> to vector<32xbf16>
     %13 = x86vector.avx512.dot %arg11, %12, %7 : vector<32xbf16> -> vector<16xf32>
     %14 = x86vector.avx512.dot %arg13, %12, %8 : vector<32xbf16> -> vector<16xf32>

     // Load the next pair of BF16 elements of A subrow, broadcast and perform BF16 dotproduct
     %15 = vector.load %sub_0[%c0, %c1, %c0, %c0] : memref<1x2x1x2xbf16, {{.*}}>, vector<2xbf16>
     %16 = vector.bitcast %15 : vector<2xbf16> to vector<1xi32>
     %17 = vector.broadcast %16 : vector<1xi32> to vector<16xi32>
     %18 = vector.bitcast %17 : vector<16xi32> to vector<32xbf16>
     %19 = x86vector.avx512.dot %arg12, %18, %7 : vector<32xbf16> -> vector<16xf32>
     %20 = x86vector.avx512.dot %arg14, %18, %8 : vector<32xbf16> -> vector<16xf32>
     
     scf.yield %13, %19, %14, %20 : vector<16xf32>, vector<16xf32>, vector<16xf32>, vector<16xf32>
    }
    scf.yield %6#0, %6#1, %6#2, %6#3 : vector<16xf32>, vector<16xf32>, vector<16xf32>, vector<16xf32>
   }

   // Store the final accumulator into C matrix
   vector.store %5#0, %subview[%c0, %c0] : memref<2x32xf32, {{.*}}>, vector<16xf32>
   vector.store %5#1, %subview[%c1, %c0] : memref<2x32xf32, {{.*}}>, vector<16xf32>
   vector.store %5#2, %subview[%c0, %c16] : memref<2x32xf32, {{.*}}>, vector<16xf32>
   vector.store %5#3, %subview[%c1, %c16] : memref<2x32xf32, {{.*}}>, vector<16xf32>
  }
 }
 return %arg2 : memref<64x64xf32>
}
\end{lstlisting}

\newpage
\subsection{BF16 VNNI packed AVX2  target}
\begin{lstlisting}[style=customC, basicstyle=\ttfamily\scriptsize, label=code:bf16-arl, caption=The AVX2 BF16 target-specific VNNI-packed nanokernel IR is generated using our proposed 
technique for the BF16 VNNI-packed input shown in Listing~\ref{code:input-bf16-vnni}. 
In the first stage - the input kernel is tiled with parameters M:2 N:16 K:2 (VNNI) - after which it is further lowered into a 
\texttt{vector.contract} operation. 
The nanokernels leverages the use of {\tt x86vector} dialect's VNNI packed operations to convert input {\tt BF16}  to {\tt FP32} type and 
does {\tt vector.fma}.]
func.func bf16(%arg0:memref<2x64x32x2xbf16>,%arg1:memref<2x32x64x2xbf16>,%arg2:memref<64x64xf32>)->memref<64x64xf32>{
 %c8 = arith.constant 8 : index
 %c0 = arith.constant 0 : index
 %c64 = arith.constant 64 : index
 %c2 = arith.constant 2 : index
 %c16 = arith.constant 16 : index
 %c1 = arith.constant 1 : index
 %c32 = arith.constant 32 : index
 scf.for %arg3 = %c0 to %c64 step %c2 { // M Loop
  scf.for %arg4 = %c0 to %c64 step %c16 { // N Loop
   %subview = memref.subview %arg2[%arg3, %arg4] [2, 16] [1, 1] : memref<64x64xf32> to memref<2x16xf32, {{.*}}>
   %1 = vector.load %subview[%c0, %c0] : memref<2x16xf32, {{.*}}>, vector<8xf32>
   %2 = vector.load %subview[%c1, %c0] : memref<2x16xf32, {{.*}}>, vector<8xf32>
   %3 = vector.load %subview[%c0, %c8] : memref<2x16xf32, {{.*}}>, vector<8xf32>
   %4 = vector.load %subview[%c1, %c8] : memref<2x16xf32, {{.*}}>, vector<8xf32>
   
   %5:4 = scf.for %arg5 = %c0 to %c2 step %c1 iter_args(%arg6 = %1, %arg7 = %2, %arg8 = %3, %arg9 = %4) -> (vector<8xf32>, vector<8xf32>, vector<8xf32>, vector<8xf32>) { // batch-reduce Loop
    %6:4 = scf.for %arg10 = %c0 to %c32 step %c1 iter_args(%arg11 = %arg6, %arg12 = %arg7, %arg13 = %arg8, %arg14 = %arg9) -> (vector<8xf32>, vector<8xf32>, vector<8xf32>, vector<8xf32>) { // K Loop
     %sub_0 = memref.subview %arg0[%arg5,%arg3,%arg10,0]{{.*}}:memref<2x64x32x2xbf16> to memref<1x2x1x2xbf16,{{.*}}>
     %sub_1 = memref.subview %arg1[%arg5,%arg10,%arg4,0]{{.*}}:memref<2x32x64x2xbf16> to memref<1x1x16x2xbf16,{{.*}}>

     // IR for the Odd FMAs.
     // Load + broadcast odd indexed element in BF16 pair of each A subrow into packed FP32 elements (emulation)
     %sub_2 = memref.subview %sub_0[0, 0, 0, 1]{{.*}}: memref<1x2x1x2xbf16, {{.*}}> to memref<1x1x1x1xbf16, {{.*}}>
     %7 = x86vector.avx.bcst_to_f32.packed %sub_2 : memref<1x1x1x1xbf16, {{.*}}> -> vector<8xf32>
     %sub_3 = memref.subview %sub_0[0, 1, 0, 1]{{.*}}: memref<1x2x1x2xbf16, {{.*}}> to memref<1x1x1x1xbf16, {{.*}}>
     %8 = x86vector.avx.bcst_to_f32.packed %sub_3 : memref<1x1x1x1xbf16, {{.*}}> -> vector<8xf32>
     // Load and convert the odd indexed element of B chunk into packed FP32 elements (emulation)
     %sub_4 = memref.subview %sub_1[0, 0 ,0, 0]{{.*}} : memref<1x1x16x2xbf16,{{.*}}> to memref<1x1x8x2xbf16,{{.*}}>
     %9 = x86vector.avx.cvt.packed.odd.indexed_to_f32 %sub_4 : memref<1x1x8x2xbf16, {{.*}}> -> vector<8xf32>
     %10 = vector.fma %7, %9, %arg11 : vector<8xf32>
     %11 = vector.fma %8, %9, %arg12 : vector<8xf32>
     %sub_5 = memref.subview %sub_1[0, 0, 8, 0]{{.*}} : memref<1x1x16x2xbf16, {{.*}}> to memref<1x1x8x2xbf16, {{.*}}>
     %12 = x86vector.avx.cvt.packed.odd.indexed_to_f32 %sub_5 : memref<1x1x8x2xbf16, {{.*}}> -> vector<8xf32>
     %13 = vector.fma %7, %12, %arg13 : vector<8xf32>
     %14 = vector.fma %8, %12, %arg14 : vector<8xf32>

     // IR for the Even FMAs. The Odd FMA acc output is used as input for Even FMA.
     // Load + broadcast even indexed element in BF16 pair of each A subrow into packed FP32 elements (emulation)
     %sub_6 = memref.subview %sub_0[0, 0, 0, 0]{{.*}}: memref<1x2x1x2xbf16, {{.*}}> to memref<1x1x1x1xbf16, {{.*}}>
     %15 = x86vector.avx.bcst_to_f32.packed %sub_6 : memref<1x1x1x1xbf16, {{.*}}> -> vector<8xf32>
     %sub_7 = memref.subview %sub_0[0, 1, 0, 0]{{.*}}: memref<1x2x1x2xbf16, {{.*}}> to memref<1x1x1x1xbf16, {{.*}}>
     %16 = x86vector.avx.bcst_to_f32.packed %sub_7 : memref<1x1x1x1xbf16, {{.*}}> -> vector<8xf32>
     %sub_8 = memref.subview %sub_1[0, 0, 0, 0]{{.*}}: memref<1x1x16x2xbf16, {{.*}}> to memref<1x1x8x2xbf16, {{.*}}>
     // Load and convert the even indexed element of B chunk into packed FP32 elements (emulation)
     %17 = x86vector.avx.cvt.packed.even.indexed_to_f32 %sub_8 : memref<1x1x8x2xbf16, {{.*}}> -> vector<8xf32>
     %18 = vector.fma %15, %17, %10 : vector<8xf32>
     %19 = vector.fma %16, %17, %11 : vector<8xf32>
     %sub_9 = memref.subview %sub_1[0, 0, 8, 0]{{.*}}: memref<1x1x16x2xbf16, {{.*}}> to memref<1x1x8x2xbf16, {{.*}}>
     %20 = x86vector.avx.cvt.packed.even.indexed_to_f32 %sub_9 : memref<1x1x8x2xbf16, {{.*}}> -> vector<8xf32>
     %21 = vector.fma %15, %20, %13 : vector<8xf32>
     %22 = vector.fma %16, %20, %14 : vector<8xf32>
     
     scf.yield %18, %19, %21, %22 : vector<8xf32>, vector<8xf32>, vector<8xf32>, vector<8xf32>
    }
    scf.yield %6#0, %6#1, %6#2, %6#3 : vector<8xf32>, vector<8xf32>, vector<8xf32>, vector<8xf32>
   }
   vector.store %5#0, %subview[%c0, %c0] : memref<2x16xf32, {{.*}}>, vector<8xf32>
   vector.store %5#1, %subview[%c1, %c0] : memref<2x16xf32, {{.*}}>, vector<8xf32>
   vector.store %5#2, %subview[%c0, %c8] : memref<2x16xf32, {{.*}}>, vector<8xf32>
   vector.store %5#3, %subview[%c1, %c8] : memref<2x16xf32, {{.*}}>, vector<8xf32>
  }
 }
 return %arg2 : memref<64x64xf32>
}
\end{lstlisting}

\newpage
\subsection{BF16 flat layout AMX target}
\begin{lstlisting}[style=customC, basicstyle=\ttfamily\scriptsize, label=code:bf16-flat, caption=Input flat BF16 batch-reduce matmul kernel without VNNI packing.]
func.func bf16(%arg0: memref<2x64x64xbf16>,%arg1:memref<2x64x64xbf16>,%arg2:memref<64x64xf32>) -> memref<64x64xf32> {
 linalg.generic {indexing_maps = [affine_map<(d0, d2, d3, d4) -> (d0, d2, d4)>, affine_map<(d0, d2, d3, d4) -> (d0, d4, d3)>, affine_map<(d0, d2, d3, d4) -> (d2, d3)>], iterator_types = ["reduction", "parallel", "parallel", "reduction"]} ins(%arg0, %arg1 : memref<2x64x64xbf16>, memref<2x64x64xbf16>) outs(%arg2 : memref<64x64xf32>) {
 ^bb0(%in: bf16, %in_1: bf16, %out: f32):
   %a = arith.extf %in : bf16 to f32
   %b = arith.extf %in_1 : bf16 to f32
   %1 = arith.mulf %a, %b : f32
   %2 = arith.addf %out, %1 : f32
   linalg.yield %2 : f32
 }
 return %arg2 : memref<64x64xf32>
}
\end{lstlisting}

\begin{lstlisting}[style=customC, basicstyle=\ttfamily\scriptsize, label=code:flat-amx, caption=The AMX target-specific generated BF16 flat-layout nanokernel IR produced using our proposed technique. In the first stage - the input kernel is tiled 
with parameters M:32 N:32 K:32 - after which it is further lowered into a 
\texttt{vector.contract} operation. To maximize performance - we employ software 
pipelining by packing the flat subpanels into a VNNI-packed layout - thereby enabling 
efficient utilization of the \texttt{AMX} tile dot-product instructions.]
func.func bf16(%arg0: memref<2x64x64xbf16>,%arg1:memref<2x64x64xbf16>,%arg2:memref<64x64xf32>) -> memref<64x64xf32> {
 %c16 = arith.constant 16 : index
 %c0 = arith.constant 0 : index
 scf.for %arg3 = %c0 to %c64 step %c32 { // M Loop
  scf.for %arg4 = %c0 to %c64 step %c32 {// N Loop
   %subview = memref.subview %arg2[%arg3, %arg4] [32, 32] [1, 1] : memref<64x64xf32> to memref<32x32xf32, {{.*}}>
   %0 = amx.tile_zero : !amx.tile<16x16xf32>
   %1 = amx.tile_zero : !amx.tile<16x16xf32>
   %2 = amx.tile_zero : !amx.tile<16x16xf32>
   %3 = amx.tile_zero : !amx.tile<16x16xf32>
   %alloca = memref.alloca() : memref<2x32x32xbf16>

   // software pipeline and interleave the first set (say S) of B subpanel into VNNI packed.
   scf.for %arg5 = %c0 to %c32 step %c2 {
    %subview_1 = memref.subview %arg1[%c0, %c0, %arg4] [1, 32, 32] [1, 1, 1] : memref<2x64x64xbf16> to memref<1x32x32xbf16, {{.*}}>
    %8 = arith.addi %c1, %arg5 : index
    %9 = arith.divui %arg5, %c2 : index
    %10 = arith.addi %c16, %9 : index
    %11 = vector.load %subview_1[%c0, %arg5, %c0] : memref<1x32x32xbf16, {{.*}}>, vector<32xbf16>
    %12 = vector.load %subview_1[%c0, %8, %c0] : memref<1x32x32xbf16, {{.*}}>, vector<32xbf16>
    %13 = vector.shuffle %11, %12 [0, 32, 1, 33, 2, 34, 3, 35, 8, 40, 9, 41, 10, 42, 11, 43, 16, 48, 17, 49, 18, 50, 19, 51, 24, 56, 25, 57, 26, 58, 27, 59] : vector<32xbf16>, vector<32xbf16>
    %14 = vector.shuffle %11, %12 [4, 36, 5, 37, 6, 38, 7, 39, 12, 44, 13, 45, 14, 46, 15, 47, 20, 52, 21, 53, 22, 54, 23, 55, 28, 60, 29, 61, 30, 62, 31, 63] : vector<32xbf16>, vector<32xbf16>
    vector.store %13, %alloca[%c0, %9, %c0] : memref<2x32x32xbf16>, vector<32xbf16>
    vector.store %14, %alloca[%c0, %10, %c0] : memref<2x32x32xbf16>, vector<32xbf16>
   }
   
   %4 = arith.subi %c2, %c1 : index
   %5 = arith.subi %c64, %c32 : index
   %6:4 = scf.for %arg5 = %c0 to %4 step %c1 iter_args(%arg6 = %0, %arg7 = %1, %arg8 = %2, %arg9 = %3) -> (!amx.tile<16x16xf32>, !amx.tile<16x16xf32>, !amx.tile<16x16xf32>, !amx.tile<16x16xf32>) { // batch-reduce Loop (excluding last iteration)
    %8:4 = scf.for %arg10 = %c0 to %5 step %c32 iter_args(%arg11 = %arg6, %arg12 = %arg7, %arg13 = %arg8, %arg14 = %arg9) -> (!amx.tile<16x16xf32>, !amx.tile<16x16xf32>, !amx.tile<16x16xf32>, !amx.tile<16x16xf32>) { // K Loop (excluding last iteration)
     %subview_1 = memref.subview %arg0[%arg5, %arg3, %arg10] [1, 32, 32] [1, 1, 1] : memref<2x64x64xbf16> to memref<1x32x32xbf16, {{.*}}>
     %10 = arith.addi %c32, %arg10 : index
     %subview_2 = memref.subview %arg1[%arg5, %10, %arg4] [1, 32, 32] [1, 1, 1] : memref<2x64x64xbf16> to memref<1x32x32xbf16, {{.*}}>

     // Load A subpanel
     %11 = amx.tile_load %subview_1[%c0, %c0, %c0] : memref<1x32x32xbf16, {{.*}}> into !amx.tile<16x32xbf16>
     %12 = amx.tile_load %subview_1[%c0, %c16, %c0] : memref<1x32x32xbf16, {{.*}}> into !amx.tile<16x32xbf16>
     
     // software pipeline and interleave the next set (S + 1) of B subpanel into VNNI packed
     scf.for %arg15 = %c0 to %c32 step %c2 {
      %21 = arith.addi %c1, %arg15 : index
      %22 = arith.divui %arg15, %c2 : index
      %23 = arith.addi %c16, %22 : index
      %24 = arith.divui %10, %c32 : index
      %25 = arith.remui %24, %c2 : index
      %26 = vector.load %subview_2[%c0, %arg15, %c0] : memref<1x32x32xbf16, {{.*}}>, vector<32xbf16>
      %27 = vector.load %subview_2[%c0, %21, %c0] : memref<1x32x32xbf16, {{.*}}>, vector<32xbf16>
      %28 = vector.shuffle %26, %27 [0, 32, 1, 33, 2, 34, 3, 35, 8, 40, 9, 41, 10, 42, 11, 43, 16, 48, 17, 49, 18, 50, 19, 51, 24, 56, 25, 57, 26, 58, 27, 59] : vector<32xbf16>, vector<32xbf16>
      %29 = vector.shuffle %26, %27 [4, 36, 5, 37, 6, 38, 7, 39, 12, 44, 13, 45, 14, 46, 15, 47, 20, 52, 21, 53, 22, 54, 23, 55, 28, 60, 29, 61, 30, 62, 31, 63] : vector<32xbf16>, vector<32xbf16>
      vector.store %28, %alloca[%25, %22, %c0] : memref<2x32x32xbf16>, vector<32xbf16>
      vector.store %29, %alloca[%25, %23, %c0] : memref<2x32x32xbf16>, vector<32xbf16>
     }
     
     %13 = arith.divui %arg10, %c32 : index
     %14 = arith.remui %13, %c2 : index
     // Load the VNNI packed (S) B subpanel 
     %15 = amx.tile_load %alloca[%14, %c0, %c0] : memref<2x32x32xbf16> into !amx.tile<16x32xbf16>
     %16 = amx.tile_load %alloca[%14, %c16, %c0] : memref<2x32x32xbf16> into !amx.tile<16x32xbf16>
     
     // tiled dot-product computation
     %17 = amx.tile_mulf %11, %15, %arg11 : !amx.tile<16x32xbf16>, !amx.tile<16x32xbf16>, !amx.tile<16x16xf32>
     %18 = amx.tile_mulf %11, %16, %arg12 : !amx.tile<16x32xbf16>, !amx.tile<16x32xbf16>, !amx.tile<16x16xf32>
     %19 = amx.tile_mulf %12, %15, %arg13 : !amx.tile<16x32xbf16>, !amx.tile<16x32xbf16>, !amx.tile<16x16xf32>
     %20 = amx.tile_mulf %12, %16, %arg14 : !amx.tile<16x32xbf16>, !amx.tile<16x32xbf16>, !amx.tile<16x16xf32>
     
     scf.yield %17, %18, %19, %20 : !amx.tile<16x16xf32>, !amx.tile<16x16xf32>, !amx.tile<16x16xf32>, !amx.tile<16x16xf32>
    }
    
    %9:4 = scf.for %arg10 = %5 to %c64 step %c32 iter_args(%arg11 = %8#0, %arg12 = %8#1, %arg13 = %8#2, %arg14 = %8#3) -> (!amx.tile<16x16xf32>, !amx.tile<16x16xf32>, !amx.tile<16x16xf32>, !amx.tile<16x16xf32>) { // K Loop (last iteration)
     %subview_1 = memref.subview %arg0[%arg5, %arg3, %arg10] [1, 32, 32] [1, 1, 1] : memref<2x64x64xbf16> to memref<1x32x32xbf16, {{.*}}>
     %10 = arith.addi %c1, %arg5 : index
     %subview_2 = memref.subview %arg1[%10, %c0, %arg4] [1, 32, 32] [1, 1, 1] : memref<2x64x64xbf16> to memref<1x32x32xbf16, {{.*}}>
     %11 = amx.tile_load %subview_1[%c0, %c0, %c0] : memref<1x32x32xbf16, {{.*}}> into !amx.tile<16x32xbf16>
     %12 = amx.tile_load %subview_1[%c0, %c16, %c0] : memref<1x32x32xbf16, {{.*}}> into !amx.tile<16x32xbf16>
     scf.for %arg15 = %c0 to %c32 step %c2 {
      %21 = arith.addi %c1, %arg15 : index
      %22 = arith.divui %arg15, %c2 : index
      %23 = arith.addi %c16, %22 : index
      %24 = vector.load %subview_2[%c0, %arg15, %c0] : memref<1x32x32xbf16, {{.*}}>, vector<32xbf16>
      %25 = vector.load %subview_2[%c0, %21, %c0] : memref<1x32x32xbf16, {{.*}}>, vector<32xbf16>
      %26 = vector.shuffle %24, %25 [0, 32, 1, 33, 2, 34, 3, 35, 8, 40, 9, 41, 10, 42, 11, 43, 16, 48, 17, 49, 18, 50, 19, 51, 24, 56, 25, 57, 26, 58, 27, 59] : vector<32xbf16>, vector<32xbf16>
      %27 = vector.shuffle %24, %25 [4, 36, 5, 37, 6, 38, 7, 39, 12, 44, 13, 45, 14, 46, 15, 47, 20, 52, 21, 53, 22, 54, 23, 55, 28, 60, 29, 61, 30, 62, 31, 63] : vector<32xbf16>, vector<32xbf16>
      vector.store %26, %alloca[%c0, %22, %c0] : memref<2x32x32xbf16>, vector<32xbf16>
      vector.store %27, %alloca[%c0, %23, %c0] : memref<2x32x32xbf16>, vector<32xbf16>
     }
     
     %13 = arith.divui %arg10, %c32 : index
     %14 = arith.remui %13, %c2 : index
     %15 = amx.tile_load %alloca[%14, %c0, %c0] : memref<2x32x32xbf16> into !amx.tile<16x32xbf16>
     %16 = amx.tile_load %alloca[%14, %c16, %c0] : memref<2x32x32xbf16> into !amx.tile<16x32xbf16>
     
     %17 = amx.tile_mulf %11, %15, %arg11 : !amx.tile<16x32xbf16>, !amx.tile<16x32xbf16>, !amx.tile<16x16xf32>
     %18 = amx.tile_mulf %11, %16, %arg12 : !amx.tile<16x32xbf16>, !amx.tile<16x32xbf16>, !amx.tile<16x16xf32>
     %19 = amx.tile_mulf %12, %15, %arg13 : !amx.tile<16x32xbf16>, !amx.tile<16x32xbf16>, !amx.tile<16x16xf32>
     %20 = amx.tile_mulf %12, %16, %arg14 : !amx.tile<16x32xbf16>, !amx.tile<16x32xbf16>, !amx.tile<16x16xf32>
     
     scf.yield %17, %18, %19, %20 : !amx.tile<16x16xf32>, !amx.tile<16x16xf32>, !amx.tile<16x16xf32>, !amx.tile<16x16xf32>
    }
    scf.yield %9#0, %9#1, %9#2, %9#3 : !amx.tile<16x16xf32>, !amx.tile<16x16xf32>, !amx.tile<16x16xf32>, !amx.tile<16x16xf32>
   }
   
   %7:4 = scf.for %arg5 = %4 to %c2 step %c1 iter_args(%arg6 = %6#0, %arg7 = %6#1, %arg8 = %6#2, %arg9 = %6#3) -> (!amx.tile<16x16xf32>, !amx.tile<16x16xf32>, !amx.tile<16x16xf32>, !amx.tile<16x16xf32>) { // batch-reduce Loop (last iteration)
    %8:4 = scf.for %arg10 = %c0 to %5 step %c32 iter_args(%arg11 = %arg6, %arg12 = %arg7, %arg13 = %arg8, %arg14 = %arg9) -> (!amx.tile<16x16xf32>, !amx.tile<16x16xf32>, !amx.tile<16x16xf32>, !amx.tile<16x16xf32>) { // K Loop (excluding last iteration)
     %subview_1 = memref.subview %arg0[%arg5, %arg3, %arg10] [1, 32, 32] [1, 1, 1] : memref<2x64x64xbf16> to memref<1x32x32xbf16, {{.*}}>
     %10 = arith.addi %c32, %arg10 : index
     %subview_2 = memref.subview %arg1[%arg5, %10, %arg4] [1, 32, 32] [1, 1, 1] : memref<2x64x64xbf16> to memref<1x32x32xbf16, {{.*}}>
     %11 = amx.tile_load %subview_1[%c0, %c0, %c0] : memref<1x32x32xbf16, {{.*}}> into !amx.tile<16x32xbf16>
     %12 = amx.tile_load %subview_1[%c0, %c16, %c0] : memref<1x32x32xbf16, {{.*}}> into !amx.tile<16x32xbf16>
     
     scf.for %arg15 = %c0 to %c32 step %c2 {
      %21 = arith.addi %c1, %arg15 : index
      %22 = arith.divui %arg15, %c2 : index
      %23 = arith.addi %c16, %22 : index
      %24 = arith.divui %10, %c32 : index
      %25 = arith.remui %24, %c2 : index
      %26 = vector.load %subview_2[%c0, %arg15, %c0] : memref<1x32x32xbf16, {{.*}}>, vector<32xbf16>
      %27 = vector.load %subview_2[%c0, %21, %c0] : memref<1x32x32xbf16, {{.*}}>, vector<32xbf16>
      %28 = vector.shuffle %26, %27 [0, 32, 1, 33, 2, 34, 3, 35, 8, 40, 9, 41, 10, 42, 11, 43, 16, 48, 17, 49, 18, 50, 19, 51, 24, 56, 25, 57, 26, 58, 27, 59] : vector<32xbf16>, vector<32xbf16>
      %29 = vector.shuffle %26, %27 [4, 36, 5, 37, 6, 38, 7, 39, 12, 44, 13, 45, 14, 46, 15, 47, 20, 52, 21, 53, 22, 54, 23, 55, 28, 60, 29, 61, 30, 62, 31, 63] : vector<32xbf16>, vector<32xbf16>
      vector.store %28, %alloca[%25, %22, %c0] : memref<2x32x32xbf16>, vector<32xbf16>
      vector.store %29, %alloca[%25, %23, %c0] : memref<2x32x32xbf16>, vector<32xbf16>
     }
     
     %13 = arith.divui %arg10, %c32 : index
     %14 = arith.remui %13, %c2 : index
     %15 = amx.tile_load %alloca[%14, %c0, %c0] : memref<2x32x32xbf16> into !amx.tile<16x32xbf16>
     %16 = amx.tile_load %alloca[%14, %c16, %c0] : memref<2x32x32xbf16> into !amx.tile<16x32xbf16>
     
     %17 = amx.tile_mulf %11, %15, %arg11 : !amx.tile<16x32xbf16>, !amx.tile<16x32xbf16>, !amx.tile<16x16xf32>
     %18 = amx.tile_mulf %11, %16, %arg12 : !amx.tile<16x32xbf16>, !amx.tile<16x32xbf16>, !amx.tile<16x16xf32>
     %19 = amx.tile_mulf %12, %15, %arg13 : !amx.tile<16x32xbf16>, !amx.tile<16x32xbf16>, !amx.tile<16x16xf32>
     %20 = amx.tile_mulf %12, %16, %arg14 : !amx.tile<16x32xbf16>, !amx.tile<16x32xbf16>, !amx.tile<16x16xf32>
     
     scf.yield %17, %18, %19, %20 : !amx.tile<16x16xf32>, !amx.tile<16x16xf32>, !amx.tile<16x16xf32>, !amx.tile<16x16xf32>
    }
    
    9:4 = scf.for %arg10 = %5 to %c64 step %c32 iter_args(%arg11 = %8#0, %arg12 = %8#1, %arg13 = %8#2, %arg14 = %8#3) -> (!amx.tile<16x16xf32>, !amx.tile<16x16xf32>, !amx.tile<16x16xf32>, !amx.tile<16x16xf32>) { // K Loop (last iteration)
     %subview_1 = memref.subview %arg0[%arg5, %arg3, %arg10] [1, 32, 32] [1, 1, 1] : memref<2x64x64xbf16> to memref<1x32x32xbf16, {{.*}}>
     %10 = amx.tile_load %subview_1[%c0, %c0, %c0] : memref<1x32x32xbf16, {{.*}}> into !amx.tile<16x32xbf16>
     %11 = amx.tile_load %subview_1[%c0, %c16, %c0] : memref<1x32x32xbf16, {{.*}}> into !amx.tile<16x32xbf16>
     %12 = arith.divui %arg10, %c32 : index
     %13 = arith.remui %12, %c2 : index
     %14 = amx.tile_load %alloca[%13, %c0, %c0] : memref<2x32x32xbf16> into !amx.tile<16x32xbf16>
     %15 = amx.tile_load %alloca[%13, %c16, %c0] : memref<2x32x32xbf16> into !amx.tile<16x32xbf16>
     
     %16 = amx.tile_mulf %10, %14, %arg11 : !amx.tile<16x32xbf16>, !amx.tile<16x32xbf16>, !amx.tile<16x16xf32>
     %17 = amx.tile_mulf %10, %15, %arg12 : !amx.tile<16x32xbf16>, !amx.tile<16x32xbf16>, !amx.tile<16x16xf32>
     %18 = amx.tile_mulf %11, %14, %arg13 : !amx.tile<16x32xbf16>, !amx.tile<16x32xbf16>, !amx.tile<16x16xf32>
     %19 = amx.tile_mulf %11, %15, %arg14 : !amx.tile<16x32xbf16>, !amx.tile<16x32xbf16>, !amx.tile<16x16xf32>
     
     scf.yield %16, %17, %18, %19 : !amx.tile<16x16xf32>, !amx.tile<16x16xf32>, !amx.tile<16x16xf32>, !amx.tile<16x16xf32>
    }
    scf.yield %9#0, %9#1, %9#2, %9#3 : !amx.tile<16x16xf32>, !amx.tile<16x16xf32>, !amx.tile<16x16xf32>, !amx.tile<16x16xf32>
   }
   
   %alloca_0 = memref.alloca() : memref<32x32xf32>
   amx.tile_store %alloca_0[%c0, %c0], %7#0 : memref<32x32xf32>, !amx.tile<16x16xf32>
   amx.tile_store %alloca_0[%c0, %c16], %7#1 : memref<32x32xf32>, !amx.tile<16x16xf32>
   amx.tile_store %alloca_0[%c16, %c0], %7#2 : memref<32x32xf32>, !amx.tile<16x16xf32>
   amx.tile_store %alloca_0[%c16, %c16], %7#3 : memref<32x32xf32>, !amx.tile<16x16xf32>
   
   // Shuffle the final accumulator and store into C matrix
   scf.for %arg5 = %c0 to %c32 step %c1 {
    %8 = vector.load %alloca_0[%arg5, %c0] : memref<32x32xf32>, vector<16xf32>
    %9 = vector.load %alloca_0[%arg5, %c16] : memref<32x32xf32>, vector<16xf32>
    %10 = vector.shuffle %8, %9 [0,1,2,3,16,17,18,19,4,5,6,7,20,21,22,23] : vector<16xf32>, vector<16xf32>
    %11 = vector.shuffle %8, %9 [8,9,10,11,24,25,26,27,12,13,14,15,28,29,30,31] : vector<16xf32>, vector<16xf32>
    %12 = vector.load %subview[%arg5, %c0] : memref<32x32xf32, {{.*}}>, vector<16xf32>
    %13 = vector.load %subview[%arg5, %c16] : memref<32x32xf32, {{.*}}>, vector<16xf32>
    %14 = arith.addf %10, %12 : vector<16xf32>
    %15 = arith.addf %11, %13 : vector<16xf32>
    vector.store %14, %subview[%arg5, %c0] : memref<32x32xf32, {{.*}}>, vector<16xf32>
    vector.store %15, %subview[%arg5, %c16] : memref<32x32xf32, {{.*}}>, vector<16xf32>
   }
  }
 }
 return %arg2 : memref<64x64xf32>
}
\end{lstlisting}

\newpage
\subsection{BF16 flat packed AVX-512 DotOp target}
\begin{lstlisting}[style=customC, basicstyle=\ttfamily\scriptsize, label=code:flat-dp, caption=The AVX-512 BF16 target-specific DotOp-based flat-layout nanokernel IR is generated 
using our proposed technique for the BF16 flat-layout input shown in 
Listing~\ref{code:bf16-flat}. In the first stage - the input kernel is tiled 
with parameters M:2 N:32 K:2 - after which it is further lowered into a 
\texttt{vector.contract} operation. The B matrix is packed into VNNI:2 format to
leverages the use of {\tt x86vector} dialect's VNNI operation ({\tt avx512.dot}).]
func.func bf16(%arg0: memref<2x64x64xbf16>,%arg1:memref<2x64x64xbf16>,%arg2:memref<64x64xf32>) -> memref<64x64xf32> {
 %c16 = arith.constant 16 : index
 %cst = arith.constant dense<0.000000e+00> : vector<16xf32>
 %c0 = arith.constant 0 : index
 ....
 scf.for %arg3 = %c0 to %c64 step %c2 { // M Loop
  scf.for %arg4 = %c0 to %c64 step %c32 { // N Loop
   %subview = memref.subview %arg2[%arg3, %arg4] [2, 32] [1, 1] : memref<64x64xf32> to memref<2x32xf32, {{.*}}>
   
   %1:4 = scf.for %arg5 = %c0 to %c2 step %c1 iter_args(%arg6 = %cst, %arg7 = %cst, %arg8 = %cst, %arg9 = %cst) -> (vector<16xf32>, vector<16xf32>, vector<16xf32>, vector<16xf32>) { // batch-reduce Loop
    %14:4 = scf.for %arg10 = %c0 to %c64 step %c2 iter_args(%arg11 = %arg6, %arg12 = %arg7, %arg13 = %arg8, %arg14 = %arg9) -> (vector<16xf32>, vector<16xf32>, vector<16xf32>, vector<16xf32>) { // K Loop
     %subview_0 = memref.subview %arg0[%arg5, %arg3, %arg10] [1, 2, 2] [1, 1, 1] : memref<2x64x64xbf16> to memref<1x2x2xbf16, {{.*}}>
     %subview_1 = memref.subview %arg1[%arg5, %arg10, %arg4] [1, 2, 32] [1, 1, 1] : memref<2x64x64xbf16> to memref<1x2x32xbf16, {{.*}}>

     // Load two chunks of B subcolumn and convert them into VNNI packed layout.
     %15 = vector.load %subview_1[%c0, %c0, %c0] : memref<1x2x32xbf16, {{.*}}>, vector<32xbf16>
     %16 = vector.load %subview_1[%c0, %c1, %c0] : memref<1x2x32xbf16, {{.*}}>, vector<32xbf16>
     %17 = vector.shuffle %15,%16 [0,32,1,33,2,34,3,35,8,40,9,41,10,42,11,43,16,48,17,49,18,50,19,51,24,56,25,57, 26,58,27,59] : vector<32xbf16>,vector<32xbf16>
     %18 = vector.shuffle %15,%16 [4,36,5,37,6,38,7,39,12,44,13,45,14,46,15,47,20,52,21,53,22,54,23,55,28,60,29,61, 30,62,31,63] : vector<32xbf16>,vector<32xbf16>

     // Load the first pair of BF16 elements of A subrow, broadcast and perform BF16 dotproduct
     %19 = vector.load %subview_0[%c0, %c0, %c0] : memref<1x2x2xbf16, {{.*}}>, vector<2xbf16>
     %20 = vector.bitcast %19 : vector<2xbf16> to vector<1xi32>
     %21 = vector.broadcast %20 : vector<1xi32> to vector<16xi32>
     %22 = vector.bitcast %21 : vector<16xi32> to vector<32xbf16>
     // Use the VNNI-packed B chunk for the dotproduct
     %23 = x86vector.avx512.dot %arg11, %22, %17 : vector<32xbf16> -> vector<16xf32>
     %24 = x86vector.avx512.dot %arg12, %22, %18 : vector<32xbf16> -> vector<16xf32>

     // Load the next pair of BF16 elements of A subrow, broadcast and perform BF16 dotproduct
     %25 = vector.load %subview_0[%c0, %c1, %c0] : memref<1x2x2xbf16, {{.*}}>, vector<2xbf16>
     %26 = vector.bitcast %25 : vector<2xbf16> to vector<1xi32>
     %27 = vector.broadcast %26 : vector<1xi32> to vector<16xi32>
     %28 = vector.bitcast %27 : vector<16xi32> to vector<32xbf16>
     %29 = x86vector.avx512.dot %arg13, %28, %17 : vector<32xbf16> -> vector<16xf32>
     %30 = x86vector.avx512.dot %arg14, %28, %18 : vector<32xbf16> -> vector<16xf32>
     
     scf.yield %23, %24, %29, %30 : vector<16xf32>, vector<16xf32>, vector<16xf32>, vector<16xf32>
    }
    scf.yield %14#0, %14#1, %14#2, %14#3 : vector<16xf32>, vector<16xf32>, vector<16xf32>, vector<16xf32>
   }
   // Shuffle the output accumulator and add the C matrix
   %2 = vector.shuffle %1#0,%1#1 [0,1,2,3,16,17,18,19,4,5,6,7,20,21,22,23] : vector<16xf32>,vector<16xf32>
   %3 = vector.shuffle %1#0,%1#1 [8,9,10,11,24,25,26,27,12,13,14,15,28,29,30,31] : vector<16xf32>,vector<16xf32>
   %4 = vector.load %subview[%c0, %c0] : memref<2x32xf32, {{.*}}>, vector<16xf32>
   %5 = arith.addf %2, %4 : vector<16xf32>
   %6 = vector.load %subview[%c0, %c16] : memref<2x32xf32, {{.*}}>, vector<16xf32>
   %7 = arith.addf %3, %6 : vector<16xf32>
   %8 = vector.shuffle %1#2,%1#3 [0,1,2,3,16,17,18,19,4,5,6,7,20,21,22,23] : vector<16xf32>,vector<16xf32>
   %9 = vector.shuffle %1#2,%1#3 [8,9,10,11,24,25,26,27,12,13,14,15,28,29,30,31] : vector<16xf32>,vector<16xf32>
   %10 = vector.load %subview[%c1, %c0] : memref<2x32xf32, {{.*}}>, vector<16xf32>
   %11 = arith.addf %8, %10 : vector<16xf32>
   %12 = vector.load %subview[%c1, %c16] : memref<2x32xf32, {{.*}}>, vector<16xf32>
   %13 = arith.addf %9, %12 : vector<16xf32>
   
   // Store the final shuffled accumulator into C matrix
   vector.store %5, %subview[%c0, %c0] : memref<2x32xf32, {{.*}}>, vector<16xf32>
   vector.store %11, %subview[%c1, %c0] : memref<2x32xf32, {{.*}}>, vector<16xf32>
   vector.store %7, %subview[%c0, %c16] : memref<2x32xf32, {{.*}}>, vector<16xf32>
   vector.store %13, %subview[%c1, %c16] : memref<2x32xf32, {{.*}}>, vector<16xf32>
  }
 }
 return %arg2 : memref<64x64xf32>
}
\end{lstlisting}

\newpage
\subsection{BF16 flat packed AVX2  target}
\begin{lstlisting}[style=customC, basicstyle=\ttfamily\scriptsize, label=code:flat-arl, caption=The AVX2 BF16 target-specific flat-layout nanokernel IR is generated using our 
proposed technique for the BF16 flat-layout input shown in 
Listing~\ref{code:bf16-flat}. In the first stage - the input kernel is tiled with parameters M:2 N:16 K:1 - after which it is further lowered into a 
\texttt{vector.contract} operation. The nanokernels leverages the use of {\tt x86vector} dialect's VNNI packed operations to convert flat input {\tt BF16}  to a VNNI packed {\tt FP32} type.]
func.func bf16(%arg0: memref<2x64x64xbf16>,%arg1:memref<2x64x64xbf16>,%arg2:memref<64x64xf32>) -> memref<64x64xf32> {
 %c8 = arith.constant 8 : index
 %cst = arith.constant dense<0.000000e+00> : vector<8xf32>
 %c0 = arith.constant 0 : index
 %c64 = arith.constant 64 : index
 %c2 = arith.constant 2 : index
 %c16 = arith.constant 16 : index
 %c1 = arith.constant 1 : index
 
 scf.for %arg3 = %c0 to %c64 step %c2 { // M Loop
  scf.for %arg4 = %c0 to %c64 step %c16 { // N Loop
   %subview = memref.subview %arg2[%arg3, %arg4] [2, 16] [1, 1] : memref<64x64xf32> to memref<2x16xf32, {{.*}}>
   %1:4 = scf.for %arg5 = %c0 to %c2 step %c1 iter_args(%arg6 = %cst, %arg7 = %cst, %arg8 = %cst, %arg9 = %cst) -> (vector<8xf32>, vector<8xf32>, vector<8xf32>, vector<8xf32>) { batch-reduce Loop
    %14:4 = scf.for %arg10 = %c0 to %c64 step %c1 iter_args(%arg11 = %arg6, %arg12 = %arg7, %arg13 = %arg8, %arg14 = %arg9) -> (vector<8xf32>, vector<8xf32>, vector<8xf32>, vector<8xf32>) { K Loop
    
     %sub_0 = memref.subview %arg0[%arg5, %arg3, %arg10] [1, 2, 1] [1,1,1] : memref<2x64x64xbf16> to memref<1x2x1xbf16, {{.*}}>
     %sub_1 = memref.subview %arg1[%arg5, %arg10, %arg4] [1, 1, 16] [1,1,1] : memref<2x64x64xbf16> to memref<1x1x16xbf16, {{.*}}>

     // Load the even and odd indexed element of B chunk and pack them into FP32 elements
     // We use the advantage of odd/even operations to load B chunk directly as VNNI packed. 
     %sub_2 = memref.subview %sub_1[0,0,0][1,1,16][1,1,1]:memref<1x1x16xbf16, {{.*}}> to memref<1x1x16xbf16, {{.*}}>
     %15 = x86vector.avx.cvt.packed.even.indexed_to_f32 %sub_2 : memref<1x1x16xbf16, {{.*}}> -> vector<8xf32>
     %sub_3 = memref.subview %sub_1[0,0,0][1,1,16][1,1,1]:memref<1x1x16xbf16, {{.*}}> to memref<1x1x16xbf16, {{.*}}>
     %16 = x86vector.avx.cvt.packed.odd.indexed_to_f32 %sub_3 : memref<1x1x16xbf16, {{.*}}> -> vector<8xf32>

     // Load + broadcast each element of A subrow into packed FP32 elements and do FMA.
     %sub_4 = memref.subview %sub_0[0,0,0] [1,1,1] [1,1,1] : memref<1x2x1xbf16, {{.*}}> to memref<1x1x1xbf16, {{.*}}>
     %17 = x86vector.avx.bcst_to_f32.packed %sub_4 : memref<1x1x1xbf16, {{.*}}> -> vector<8xf32>
     %18 = vector.fma %17, %15, %arg11 : vector<8xf32>
     %19 = vector.fma %17, %16, %arg12 : vector<8xf32>
     %sub_5 = memref.subview %sub_0[0,1,0] [1,1,1] [1,1,1] : memref<1x2x1xbf16, {{.*}}> to memref<1x1x1xbf16, {{.*}}>
     %20 = x86vector.avx.bcst_to_f32.packed %sub_5 : memref<1x1x1xbf16, {{.*}}> -> vector<8xf32>
     %21 = vector.fma %20, %15, %arg13 : vector<8xf32>
     %22 = vector.fma %20, %16, %arg14 : vector<8xf32>
     
     scf.yield %18, %19, %21, %22 : vector<8xf32>, vector<8xf32>, vector<8xf32>, vector<8xf32>
    }
    scf.yield %14#0, %14#1, %14#2, %14#3 : vector<8xf32>, vector<8xf32>, vector<8xf32>, vector<8xf32>
   }
   // Shuffle the output accumulator and add the C matrix
   %2 = vector.shuffle %1#0, %1#1 [0, 8, 1, 9, 2, 10, 3, 11] : vector<8xf32>, vector<8xf32>
   %3 = vector.shuffle %1#0, %1#1 [4, 12, 5, 13, 6, 14, 7, 15] : vector<8xf32>, vector<8xf32>
   %4 = vector.load %subview[%c0, %c0] : memref<2x16xf32, {{.*}}>, vector<8xf32>
   %5 = arith.addf %2, %4 : vector<8xf32>
   %6 = vector.load %subview[%c0, %c8] : memref<2x16xf32, {{.*}}>, vector<8xf32>
   %7 = arith.addf %3, %6 : vector<8xf32>
   %8 = vector.shuffle %1#2, %1#3 [0, 8, 1, 9, 2, 10, 3, 11] : vector<8xf32>, vector<8xf32>
   %9 = vector.shuffle %1#2, %1#3 [4, 12, 5, 13, 6, 14, 7, 15] : vector<8xf32>, vector<8xf32>
   %10 = vector.load %subview[%c1, %c0] : memref<2x16xf32, {{.*}}>, vector<8xf32>
   %11 = arith.addf %8, %10 : vector<8xf32>
   %12 = vector.load %subview[%c1, %c8] : memref<2x16xf32, {{.*}}>, vector<8xf32>
   %13 = arith.addf %9, %12 : vector<8xf32>
   
   // Store the final shuffled accumulator into C matrix
   vector.store %5, %subview[%c0, %c0] : memref<2x16xf32, {{.*}}>, vector<8xf32>
   vector.store %11, %subview[%c1, %c0] : memref<2x16xf32, {{.*}}>, vector<8xf32>
   vector.store %7, %subview[%c0, %c8] : memref<2x16xf32, {{.*}}>, vector<8xf32>
   vector.store %13, %subview[%c1, %c8] : memref<2x16xf32, {{.*}}>, vector<8xf32>
  }
 }
 return %arg2 : memref<64x64xf32>
}
\end{lstlisting}







\end{document}